\documentclass[journal]{IEEEtran}
\usepackage{amsmath,  amsfonts}
\usepackage{algorithmic}
\usepackage{algorithm}
\usepackage{array}
\usepackage[caption=false,  font=normalsize,  labelfont=sf,  textfont=sf]{subfig}
\usepackage{textcomp}
\usepackage{stfloats}
\usepackage{url}
\usepackage{verbatim}
\usepackage{graphicx}
\usepackage{booktabs}
\usepackage{color}
\usepackage{cite}
\usepackage{multirow}
\usepackage{bm}
\usepackage{overpic}

\usepackage[colorlinks,
	linkcolor=blue,       %
	anchorcolor=blue,  %
	citecolor=blue,        %
]{hyperref}

\newcommand{\tabincell}[2]{\begin{tabular}{@{}#1@{}}#2\end{tabular}}
\begin{document}
\setcounter{secnumdepth}{4} %
\setcounter{tocdepth}{4} %

\title{From Static to Dynamic: \\Adapting Landmark-Aware Image Models\\  for Facial Expression Recognition in Videos}

\author{Yin~Chen$^{\dagger}$, Jia~Li$^{\dagger \ast}$, Shiguang~Shan,~\IEEEmembership{Fellow,~IEEE,}

	Meng~Wang,~\IEEEmembership{Fellow,~IEEE,} and
	Richang~Hong,~\IEEEmembership{Member,~IEEE}

	\thanks{
		$\dagger$ Equal contribution. $\ast$ Corresponding author.
		
		This work was supported in part by the National Key Research and Development Program of China under Grant 2019YFA0706203 and also in part by the National Natural Science Foundation of China under Grant  62202139, and also in part by The University Synergy Innovation Program of Anhui Province (GXXT-2022-038).  

		Yin Chen, Jia Li, Meng Wang and Richang Hong are with the School of Computer
		Science and Information Engineering, Hefei University of Technology, Hefei
		230601, China (e-mail: chenyin@mail.hfut.edu.cn; jiali@hfut.edu.cn;
		eric.mengwang@gmail.com; hongrc.hfut@gmail.com).

		Shiguang Shan is with the Key Laboratory of Intelligent Information Processing,
		Institute of Computing Technology, Chinese Academy of Sciences, Beijing 100190,
		China, and also with the University of Chinese Academy of Sciences, Beijing,
		100049, China (e-mail: sgshan@ict.ac.cn).

		© 2024 IEEE. Personal use of this material is permitted. Permission from IEEE must be obtained for all other uses, in any current or future media, including reprinting/republishing this material for advertising or promotional purposes, creating new collective works, for resale or redistribution to servers or lists, or reuse of any copyrighted component of this work in other works.
		Digital Object Identifier: \href{https://doi.org/10.1109/TAFFC.2024.3453443}{10.1109/TAFFC.2024.3453443}
		}
}

\maketitle

\begin{abstract}
	Dynamic facial expression recognition (DFER) in the wild is still hindered by data limitations, e.g., insufficient quantity and diversity of pose, occlusion and illumination, as well as the inherent ambiguity of facial expressions. In contrast, static facial expression recognition (SFER) currently shows much higher performance and can benefit from more abundant high-quality training data. Moreover, the appearance features and dynamic dependencies of DFER remain largely unexplored. Recognizing the potential in leveraging SFER knowledge for DFER, we introduce a novel Static-to-Dynamic model (S2D) that leverages existing SFER knowledge and dynamic information implicitly encoded in extracted facial landmark-aware features, thereby significantly improving DFER performance. Firstly, we build and train an image model for SFER, which incorporates a standard Vision Transformer (ViT) and Multi-View Complementary Prompters (MCPs) only. Then, we obtain our video model (i.e., S2D), for DFER, by inserting Temporal-Modeling Adapters (TMAs) into the image model. MCPs enhance facial expression features with landmark-aware features inferred by an off-the-shelf facial landmark detector. And the TMAs capture and model the relationships of dynamic changes in facial expressions, effectively extending the pre-trained image model for videos. Notably, MCPs and TMAs only increase a fraction of trainable parameters (less than +10\%) to the original image model. Moreover, we present a novel Emotion-Anchors (i.e., reference samples for each emotion category) based Self-Distillation Loss to reduce the detrimental influence of ambiguous emotion labels, further enhancing our S2D. Experiments conducted on popular SFER and DFER datasets show that we have achieved a new state of the art.
\end{abstract}


\section{Introduction}
\IEEEPARstart{F}{acial} expressions reflect a person's emotional state and play a crucial role in interpersonal interactions. Understanding the emotional states from facial expressions is increasingly significant due to its applications, such as human-computer interaction \cite{Liu2017AFE}, healthcare aids \cite{Bisogni2022ImpactOD}, and driving safety \cite{Wilhelm2019TowardsFE}. Currently, facial expression recognition (FER) can be roughly divided into two types: Static Facial Expression Recognition (SFER) and Dynamic Facial Expression Recognition (DFER). SFER mainly focuses on recognizing expressions from static images, whereas DFER concentrates on recognizing expressions from dynamic image sequences (or videos).

\begin{figure}
	\centering
	\begin{overpic}[width=0.45\textwidth]{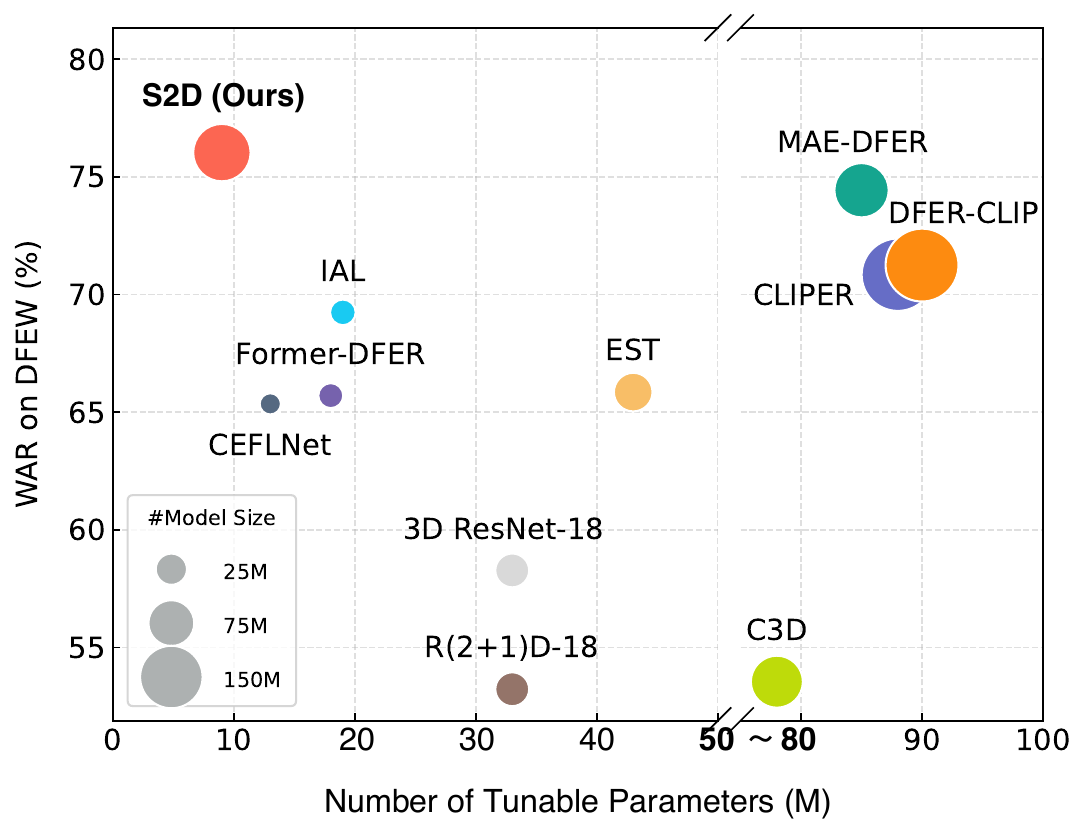}
	\end{overpic}
	\caption{Performance comparison of dynamic facial expression recognition on the DFEW \cite{jiang2020dfew} testing set. Bubble size indicates the model size. Our proposed S2D achieves the highest weighted average recall (WAR) while enjoying significantly less number of tunable parameters ($<10\%$ tunable parameters of the whole model). Here, we compare our S2D with C3D \cite{Tran2014LearningSF}, R(2+1)D-18  \cite{tran2018closer}, 3D ResNet-18 \cite{He2015DeepRL}, Former-DFER \cite{Zhao2021FormerDFERDF}, CEFLNet \cite{liu2022clip},  EST \cite{liu2023expression}, IAL \cite{li2023intensity}, CLIPER \cite{Li2023CLIPERAU}, DFER-CLIP \cite{zhao2023prompting} and MAE-DFER \cite{sun2023mae}.}
	\label{fig:params}
\end{figure}
With the advent of deep learning, FER has made considerable progress on real-world SFER datasets (e.g., RAF-DB \cite{Li2017ReliableCA}, AffectNet \cite{Mollahosseini2017AffectNetAD}, FERPlus \cite{Barsoum2016TrainingDN}), but the performance on in-the-wild DFER datasets (e.g., DFEW \cite{jiang2020dfew}, FERV39K \cite{wang2022ferv39K}, MAFW \cite{Liu2022MAFWAL}) is still far from satisfactory. This is primarily due to the difficulty in collecting DFER data, the limitations of dataset representativeness (e.g., poses, occlusions, and illuminations), the inherent ambiguity of facial expressions, and the insufficient mining of temporal information.

\textbf{How to reduce the negative impact of data deficiencies and ambiguous annotations remains a challenge for DFER.}  Publicly available large-scale datasets for SFER are in-the-wild images, mainly crawled from the Internet, and DFER datasets are commonly constructed by collecting videos from movies, TV dramas, and various online sources. 
Regardless of differences in data sources and quality, there is a noticeable shortage of video data compared to image data, creating a substantial disparity in sample and subject quantity between SFER and DFER datasets. To illustrate, the currently largest DFER dataset, FERV39K \cite{wang2022ferv39K}, comprises just 38,935 samples, while the largest SFER dataset AffectNet \cite{Mollahosseini2017AffectNetAD} contains approximately 450,000 manually labeled samples. Consequently, data-driven methods face more pronounced limitations in the performance of DFER. Additionally, we have conducted an inter-annotator agreement analysis on the FERPlus \cite{Barsoum2016TrainingDN} and DFEW \cite{jiang2020dfew} datasets using the Fleiss’ Kappa \cite{fleiss1971measuring} metric. The results show that the inter-annotator agreement on the DFEW dataset (0.50) is lower than that on the FERPlus dataset (0.55). This finding highlights the higher complexity of facial expressions in video data and the variability among annotators, which collectively contribute to a slightly higher level of ambiguity in DFER datasets compared to SFER datasets. In summary, the DFER task exhibits deficiencies in both quantity and annotation quality in contrast to SFER, which \emph{poses a formidable challenge for enhancing DFER performance solely relying on limited DFER data.} Considering that collecting a sufficiently large DFER dataset is also challenging, requiring huge labor and financial resources, it's too difficult for us to reconstruct an extensive DFER dataset. However, the existing SFER data are abundant and diverse enough, and share significant similar knowledge to the DFER data. So why not leverage SFER data as prior knowledge to improve the performance of DFER?

\textbf{Another challenge for DFER lies in how to effectively model the temporal dimension to capture the dynamic features of facial expressions in videos.} Compared to a static image, a video clip contains significant temporal information. Viewing emotional state as a dynamic process, an individual's emotional state changes over time. Hence, accurately capturing the dynamic changes of emotions is crucial for achieving high performance in DFER modeling. Researchers have introduced various methods to address this, with promising results. These methods include the utilization of techniques such as 3D Convolutional Neural Networks \cite{Fan2016VideobasedER}, 2D Convolutional Neural Networks \cite{jiang2020dfew, Kossaifi2019FactorizedHC} combined with Recurrent Neural Networks \cite{Sun2020MultimodalCD}, and CNN combined with Transformer architectures \cite{Zhao2021FormerDFERDF,  Ma2022SpatioTemporalTF,  li2023intensity,  li2023multimodal}. However, while some recent studies, such as MARLIN \cite{Cai_2023_CVPR} and MAE-DFER\cite{sun2023mae}, have begun to address the temporal dynamics of facial expressions through advanced self-supervised representation learning like mask modeling, the majority of existing methods in DFER still primarily focus on implicitly learning the temporal relationships of dynamic facial expressions through end-to-end training, which map video clips directly to category labels.  These methods often rely on relatively sparse supervised signals and do not explicitly model the dynamic changes of expressive faces throughout the video, thereby limiting their ability to fully capture the temporal dynamics of facial expressions.

To mitigate the above main challenges, novel methods that can harness prior
knowledge, efficiently capture dynamic emotional changes in videos, and provide
more reliable supervision signals should be explored to enhance DFER
performance. CLIPER \cite{Li2023CLIPERAU} and DFER-CLIP
\cite{zhao2023prompting} have leveraged prior knowledge from the very powerful
vision-language model CLIP \cite{radford2021learning} to enhance DFER,
achieving promising results. However, these vision-language models still
underperform the current state-of-the-art visual-only method MAE-DFER
\cite{sun2023mae}, despite leveraging vision and language knowledge from CLIP
simultaneously. The I3D model \cite{carreira2017quo} can transfer vision
knowledge learned from static images to videos, improving performance in
downstream tasks. Nevertheless, models like I3D require retraining their all
parameters to capture more fine-grained spatiotemporal features, leading to
high computational costs. In contrast, the adapter-based learning approaches
\cite{chen2022adaptformer, Yang2023AIMAI, pan2022st, Chen_2023_ICCV} can
transfer an image model to videos efficiently by fine-tuning only a subset of
parameters. However, due to the absence of prior facial knowledge, there exists
a notable domain gap between the general features of these upstream models and
the specific FER features, making direct application to FER unfeasible. To
bridge this gap, researchers have integrated Action Units (AU)
\cite{liu2013aware, chen2021understanding, liang2020fine} and facial landmarks
\cite{Jung2015JointFI, Belmonte9599570, Chen7518582, zheng2023poster} as prior
facial knowledge into FER. Facial landmark detection technology has attained a
notable level of maturity, particularly in challenging real-world scenarios,
making it more suitable for in-the-wild FER than AU. Moreover, we have observed
and validated that static features learned on AffectNet dataset
\cite{Mollahosseini2017AffectNetAD} exhibit strong robustness and are more
appropriate for related downstream tasks \cite{Li_2023_CVPR}, which
significantly contributed to our victory in winning the championship of
Emotional Reaction Intensity Estimation Challenge held by the CVPR2023-ABAW5
\cite{Kollias_2023_CVPR} competition. Additionally, existing facial
landmark-based methods still lack simplicity. For example, Poster
\cite{zheng2023poster} employs multi-level landmark features, making the method
progressively complex. This motivates us to seek a simple and effective
structure.

Building upon the aforementioned insights, we propose a simple yet powerful
model, named S2D, which is designed to expand the Static FER model to the
Dynamic FER task without retraining all model parameters. Notably, S2D achieves
a new state-of-the-art performance with only a fraction of tunable parameters
(less than 10\% of the whole model parameters), as shown in Fig.
\ref{fig:params}. Specifically, S2D is a standard Vision Transformer (ViT) that
incorporates Multi-View Complementary Prompter (MCP) and Temporal-Modeling
Adapter (TMA) modules. MCP is a simple module that combines static facial
expression features and landmark-aware features, thereby enhancing the
image-level representational ability for both SFER and DFER tasks. TMA consists
of a temporal adapter with Temporal Multi-Headed Self-Attention (T-MSA)
\cite{Yang2023AIMAI} and a Vanilla Adapter \cite{houlsby2019parameter}. This
configuration empowers TMA to efficiently capture the dynamic changes of facial
expressions in the temporal dimension and transfer the learned representation
knowledge from SFER to DFER. Furthermore, we propose an Emotion-Anchors (i.e.,
base reference expressions) based Self-Distillation Loss (SDL) to provide a
reliable auxiliary supervision signal and help model learning. We performed
experiments on widely used SFER datasets (RAF-DB \cite{Li2017ReliableCA},
AffectNet \cite{Mollahosseini2017AffectNetAD}, FERPlus
\cite{Barsoum2016TrainingDN}) and DFER datasets (DFEW \cite{jiang2020dfew},
FERV39K \cite{wang2022ferv39K}, MAFW \cite{Liu2022MAFWAL}). Experimental
results demonstrate that our proposed method achieved superior or comparable
recognition accuracy compared to the state-of-the-art methods, confirming the
efficacy of our approach in learning and transferring facial expression
representations. The code and model are publicly available
here\footnote{\url{https://github.com/MSA-LMC/S2D}}.

Overall, the main contributions of this work can be summarized as follows:
\begin{itemize}
	\item[$\bullet$]
		\textbf{Image-level representation enhancement.}
		We introduce facial landmark-aware features extracted by MobileFaceNet \cite{Chen2018MobileFaceNetsEC} as another view of the raw input, which are sensitive to key facial regions, to guide the model to focus more on the emotion-related facial areas. Then we utilize both the static features learned on AffectNet \cite{Mollahosseini2017AffectNetAD} and facial landmark-aware features as prior knowledge for SFER and DFER tasks. By fusing these two features with the Multi-View Complementary Prompter (MCP), the image-level representation is significantly enhanced. It is also helpful for the DFER model to effectively capture the dynamic changes of expressions, for the facial dynamic information (e.g., muscle movements) is implicitly encoded in the sequential landmark-aware features.
	\item[$\bullet$]
		\textbf{Expanding the static FER model to the dynamic FER model efficiently.}
		We propose a Temporal-Modeling Adapter (TMA) module to efficiently expand the static FER model to the dynamic FER model by separate temporal modeling. With this ability, our FER model achieves SOTA performance on various DFER benchmarks, while maintaining tremendously parameter-efficient (only tuning $<10\%$ parameters of the whole model). By offering a simple yet powerful baseline, our method provides a user-friendly solution for the research community, which is both easy to implement and follow.
	\item[$\bullet$]
		\textbf{Emotion-Anchors based Self-Distillation Loss.} Within this loss, we utilize a group of reference samples to generate more reliable soft labels with probability distributions on all emotion classes. Such auxiliary supervision provided by this loss prevents the ambiguous emotion labels from deteriorating the FER model, further improving our DFER model.
\end{itemize}

\section{Related Work}
This paper introduces a novel method to transfer the pre-trained image model
for DFER efficiently. Therefore, we mainly review the prior works about DFER
and efficient transfer learning techniques.

\subsection{Dynamic Facial Expression Recognition in the Wild}
In the early stages, researchers conducted evaluations on datasets in
laboratory settings, relying on manually designed features. As deep learning
methods gained prominence and large-scale DFER datasets became available in
recent years, researchers started employing data-driven techniques to tackle
in-the-wild DFER challenges. At present, DFER research methods fall into three
main categories. The first category involves using 3D CNN
\cite{Fan2016VideobasedER} to simultaneously model temporal and spatial
information. However, this method is computationally expensive, and the models
may not be easily scalable to deeper architectures. The second combines 2D CNN
with RNN \cite{jiang2020dfew, Kossaifi2019FactorizedHC, Sun2020MultimodalCD}.
It first extracts features from each frame using 2D CNN and then models
temporal information using RNN. The third emerging trend in research involves
the utilization of Transformer methods. The research conducted by Former-DFER
\cite{Zhao2021FormerDFERDF} utilizes convolution-based spatial Transformer and
temporal Transformer to achieve spatiotemporal fusion. STT
\cite{Ma2022SpatioTemporalTF} employs ResNet18 to extract features and, in
combination with Transformer, jointly learns spatiotemporal features. IAL
\cite{li2023intensity} introduces a global convolution-attention block and an
intensity-aware loss to differentiate samples based on varying expression
intensities. In contrast to the aforementioned approaches, our method leverages
the prior knowledge of facial landmark detections
\cite{Chen2018MobileFaceNetsEC} and SFER data.
\begin{figure*}
	\centering
	\begin{overpic}[width=1\textwidth]{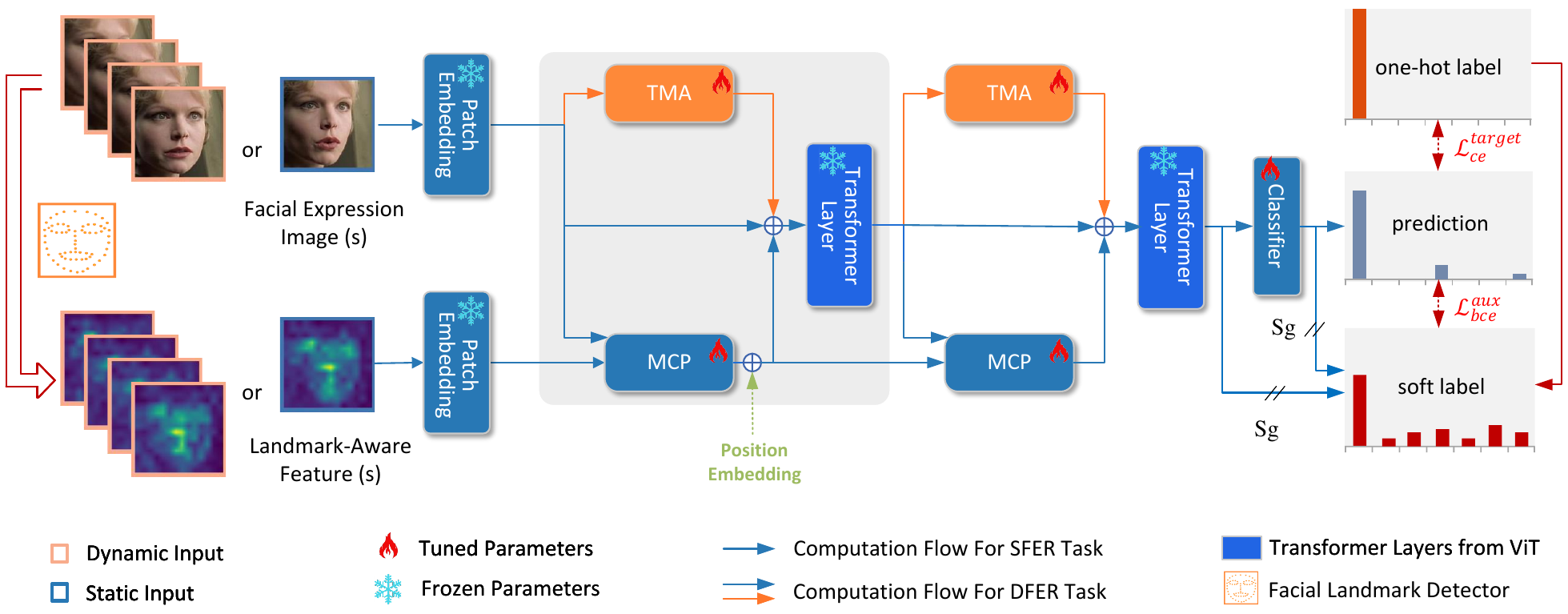}
		\put(31.7,  32){\small $\bm{\mathcal{H}}^0$}
		\put(47,  33.5){\small $\bm{\mathcal{T}}^l$}
		\put(31.7,  17){\small $\bm{\mathcal{A}}^0$}
		\put(46.8,  17){\small $\bm{\mathcal{P}}^l$}
		\put(54,  30.2){\small $E^l$}
		\put(49.5,  36.5){\small $\times(L - 1)$}
	\end{overpic}
	\caption{\textbf{Overall architecture of the proposed method.} Our S2D accepts as input a facial expression image (or facial expression image sequence) $\bm{X}_F$ and a landmark-aware feature (or landmark-aware feature sequence) $\bm{X}_L$. The facial expression image and landmark-aware feature are embedded with patch embedding layers and fed into the transformer layers $\{E^l\}^{L-1}_{l=0}$ borrowed from ViT. Temporal-Modeling Adapter (TMA) is used to capture temporal information $\bm{\mathcal{T}}^l$ while Multi-View Complementary Prompter (MCP) uses landmark-aware features to generate guiding prompts $\bm{\mathcal{P}}^l$ to enhance the image-level representational ability for both SFER and DFER tasks. Note that the position embedding is added to $\bm{\mathcal{P}}^0$ after the first MCP block, and TMA is only used for the DFER task. Sg means stop gradient.}
	\label{fig:network}
\end{figure*}
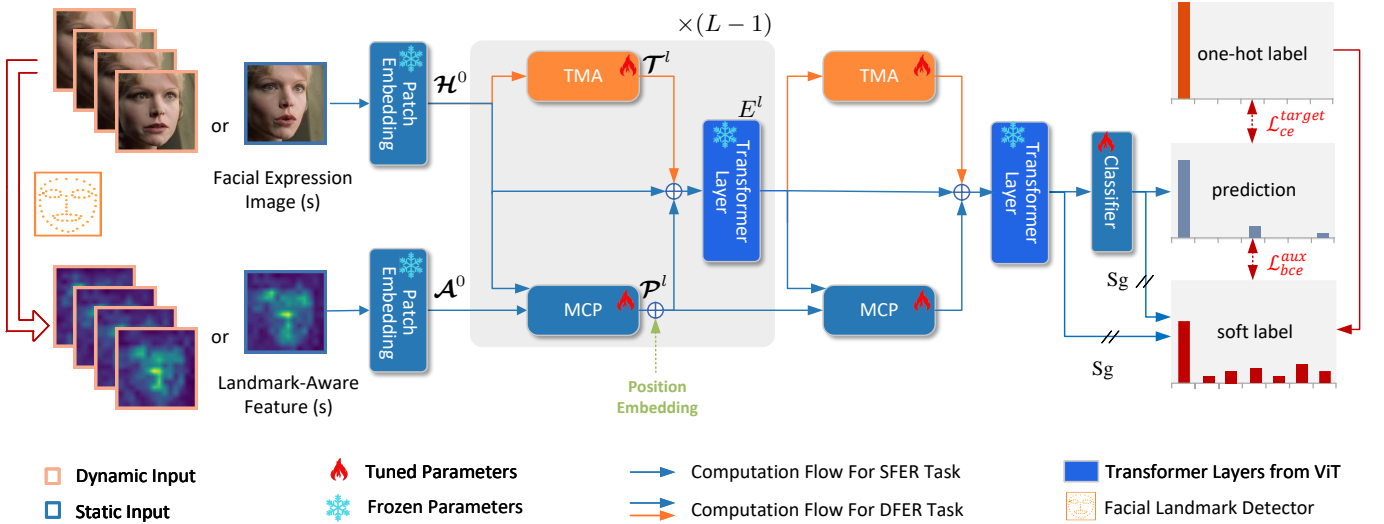

\subsection{Parameter-Efficient Transfer Learning}
Parameter-efficient transfer learning techniques \cite{houlsby2019parameter,
	hu2021lora, lester2021power, he2021towards, qing2023mar} are first introduced
in natural language processing (NLP) due to the rising computation costs of
fully fine-tuning the expanding language models for various downstream tasks.
The goal of these techniques is to decrease the number of trainable parameters
thus reducing the computational costs, while achieving a comparable or better
performance compared to full fine-tuning. Recently, parameter-efficient
transfer learning is also introduced in computer vision, and it can be divided
into prompt \cite{jia2022visual, zhu2023visual,sohn2023visual, bar2022visual}
learning and adapting \cite{chen2022adaptformer, Yang2023AIMAI, pan2022st,
	Chen_2023_ICCV} learning techniques. The former usually focuses on transfer
learning in the same domain (e.g., image-to-image or video-to-video), whereas
the latter often targets adapting an image model for video-based tasks. The
work most related to ours is AIM \cite{Yang2023AIMAI}. It utilizes lightweight
adapters to modify the vision transformer blocks and adapt a frozen,
pre-trained image model for performing video action recognition. However, there
are several major differences. Firstly, AIM adapts to downstream tasks by
modifying the original transformer layer directly, while our TMA is placed
between the adjacent transformer blocks in the form of residual. Secondly, AIM
uses an unlearnable self-attention for temporal modeling while our TMA employs
a learnable one.

\section{METHODOLOGY}
This section elaborates on our novel Static-to-Dynamic model (S2D). Rather than
devising and fully fine-tuning an existing video model or even training a DFER
model from scratch, S2D adapts a landmark-aware image model for facial
expression recognition in videos (i.e., DFER) effectively and efficiently. To
be specific, we first train a SFER model using static FER datasets, and then
merely add and tune a small fraction of adaptation-learning parameters to
achieve effective transfer learning for the DFER task. This unique approach
yields promising performance on DFER benchmarks while simultaneously ensuring
parameter efficiency.

An overall of our model is presented in Fig. \ref{fig:network}, showing our
model architecture, computation flow (detailed in section \ref{section:pre}),
multi-view input and interaction (detailed in section \ref{section:mcp}), and
training loss (detailed in section \ref{section:loss}). Section
\ref{section:tma} introduces how to expand the image model to an efficient
video model in detail.
\subsection{Preliminary}
\label{section:pre}
\subsubsection{Problem Formulation}
Given a video clip or a facial image sequence $\bm{X}_F$ with an emotion label
$\bm{Y}$, the task of DFER is to learn a mapping function $\mathcal{F}_\theta
	(\bm{X}_F) \rightarrow \bm{Y}$, where $\mathcal{F}_\theta$ denotes a model and
$\bm{\theta}$ represents its learnable parameters. In this work, we employ the
facial landmark-aware features $\bm{X}_L$ extracted from $\bm{X}_F$ using an
off-the-shelf facial landmark detector as an auxiliary view. Thus, the input of
$\mathcal{F}_\theta$ is extended to $(\bm{X}_F, \bm{X}_L)$. Accordingly, the
mapping goal has changed to $\mathcal{F}_\theta (\bm{X}_F, \bm{X}_L)
	\rightarrow \bm{Y}$.
\subsubsection{S2D Model}
As illustrated in Fig. \ref{fig:network}, our devised S2D accepts as input the
facial image sequence $\bm{X}_F\in \mathbb{R}^{T\times C \times H\times W}$ and
facial landmark-aware feature sequence $\bm{X}_L \in \mathbb{R}^{T\times C'
		\times H'\times W'}$. Here, $T$, $C$, $H$, and $W$ represent the number of
frames, channels, width, and height of the facial image sequence, respectively.
Firstly, we feed the two flows, $\bm{X}_F$ and $\bm{X}_L$, into individual
patch embedding layers separately. Each input frame is divided into $N$
patches, then mapped and flattened into a $D$-dimensional latent space. We
refer to these embeddings as the facial expression tokens $\bm{\mathcal{H}}^0
	\in \mathbb{R}^{T\times N \times D}$ and facial landmark tokens $\
	\bm{\mathcal{A}}^0 \in \mathbb{R}^{T\times N \times D}$. Then,
$\bm{\mathcal{H}}^0$ and a $[class]$ token $\bm{x}_{class}$ with position
embeddings are fed into the transformer layers $\{E^l\}_{l=0}^{L-1}$ of Vision
Transformer (ViT) \cite{Dosovitskiy2020AnII}. Note that $\bm{\mathcal{H}}^0$
and $\bm{\mathcal{A}}^0$ are firstly sent into the MCP module to generate
guiding prompts, while $\bm{\mathcal{H}}^0$ is sent to the TMA to capture the
temporal information. Finally, the learned guiding prompts $\bm{\mathcal{P}}\in
	\mathbb{R}^{T\times N \times D} $ and temporal information tokens
$\bm{\mathcal{T}} \in \mathbb{R}^{T\times N \times D} $ are added to the
original facial expression tokens in a form of residual:
\begin{equation}
	\bm{\mathcal{H}}^{l'} = \bm{\mathcal{H}}^{l} + \bm{\mathcal{P}}^{l+1}+\bm{\mathcal{T}}^{l+1}, \quad l={0,  1,  \cdots,  L-1,}
\end{equation}
where $\bm{\mathcal{H}}^{l'}$ is the prompted tokens, and $\bm{\mathcal{P}}^{l+1}$, $\bm{\mathcal{T}}^{l+1}$ are guiding prompts and temporal information from the ($l+1$)-th MCP module and TMA module, respectively.
Next, $\bm{\mathcal{H}}^{l'}$ is fed into the transformer layer $E^l$ to extract a more powerful image-level representation:
\begin{equation}
	\bm{\mathcal{H}}^{l+1} = E^l(\bm{x}_{class}, \bm{\mathcal{H}}^{l'}).
\end{equation}

\subsection{Image-Level Representation Enhancement}
\label{section:mcp}
In this section, we will introduce how to enhance image-level representational ability through the selection of static facial expression features, the incorporation of facial landmark-aware features, and the generation of guiding prompts.
\subsubsection{Selection of Static Facial Expression Feature}
To harness the image-level representational capacity of the SFER model and gain
more robust static facial expression features, we employ AffectNet
\cite{Mollahosseini2017AffectNetAD} as our pre-training dataset. AffectNet is
the existing largest SFER dataset, which contains more than 1M face images from
the Internet and about 450,000 manually annotated images. We first train the
model on the AffectNet dataset to get a robust expression appearance
representation, then fine-tune it on other FER datasets.

\subsubsection{Facial Landmark-Aware Feature}
To bolster image-level representation for both SFER and DFER, we introduce
facial landmark-aware features extracted from facial images as an auxiliary
view to guide model learning. These facial landmark-aware features are the
features in the penultimate stage of landmark detector MobileFaceNet
\cite{Chen2018MobileFaceNetsEC}. And they exhibit sensitivity to key facial
regions, such as the mouth, nose, and eyes, enabling the model to focus more on
local details relevant to facial expressions. They are used to enhance the
model's capacity to represent in-the-wild facial expressions. Furthermore, the
process of dynamic facial changes (e.g., muscle movements) is implicitly
encoded within the sequence of facial landmark-aware features, which is also
used to enhance the model's ability to capture dynamic information of facial
expressions in videos.

\subsubsection{Guiding Prompts Generation}
To fully exploit the potential of both facial expression and facial
landmark-aware features, we utilize a Multi-View Complementary Prompter (MCP)
module to generate guiding prompts, which is proposed in the field of
multi-modal object tracking and originally named Modality-Complementary
Prompter in \cite{zhu2023visual}. MCP takes in facial expression tokens
represented as $\bm{\mathcal{H}}^l$ and facial landmark tokens represented as
$\bm{\mathcal{A}}^l$, producing guiding prompts $\bm{\mathcal{P}}^{l+1}$ for
the next transformer layer. The process can be described as follows:

Firstly, $\bm{\mathcal{H}}^l$ and $\bm{\mathcal{A}}^l$ are reshaped and
projected to a lower-dimensional latent space:
\begin{equation}
	\bm{\mathcal{M}}_\mathcal{H}=g_1(\bm{\mathcal{H}}^l),\qquad \bm{\mathcal{M}}_\mathcal{A}=g_2(\bm{\mathcal{A}}^l),
\end{equation}
where $g_1$ and $g_2$ represent simple $1\times1$ convolutional layers, $\bm{\mathcal{M}}_\mathcal{H} \in \mathbb{R}^{T \times D' \times \sqrt{N} \times \sqrt{N}}$ and $\bm{\mathcal{M}}_\mathcal{A}\in \mathbb{R}^{T \times D' \times \sqrt{N} \times \sqrt{N}}$ are projected embeddings. Here, $D'$ is the output channel number of the convolutional layer. Subsequently, the spatial attention-like operation is performed on $\bm{\mathcal{M}}_\mathcal{H}$ to emphasize details related to facial expressions:
\begin{equation}
	\bm{\mathcal{M}}_{fovea} = \left\{\frac{e^{\bm{\mathcal{M}}_\mathcal{H}{[:,:,  i,  j]}}}{\sum e^{\bm{\mathcal{M}}_\mathcal{H}{[:,:,  i,  j]}} }\lambda \right\},
\end{equation}
\begin{equation}
	\bm{\mathcal{M}}^e_\mathcal{H} = \bm{\mathcal{M}}_\mathcal{H} \odot \bm{\mathcal{M}}_{fovea},
\end{equation}
where $i=1,  2,  \cdots, \sqrt{N}$, $j=1,  2,  \cdots, \sqrt{N}$,  and $\lambda$ is a learnable weighted parameter per block \cite{zhu2023visual}. Finally,  the learned guiding prompts are obtained as follows:
\begin{equation}
	\bm{\mathcal{P}}^{l+1} = g_3(\bm{\mathcal{M}}^e_H+\bm{\mathcal{M}}_A),
\end{equation}
where $g_3$ is a $1\times1$ convolutional layer, and $\bm{\mathcal{P}}^{l+1} \in \mathbb{R}^{T \times N \times D}$.
\subsection{Expanding the Image Model to Efficient Video Model}
\label{section:tma}
\begin{figure}
	\centering
	\begin{overpic}[width=0.3\textwidth]{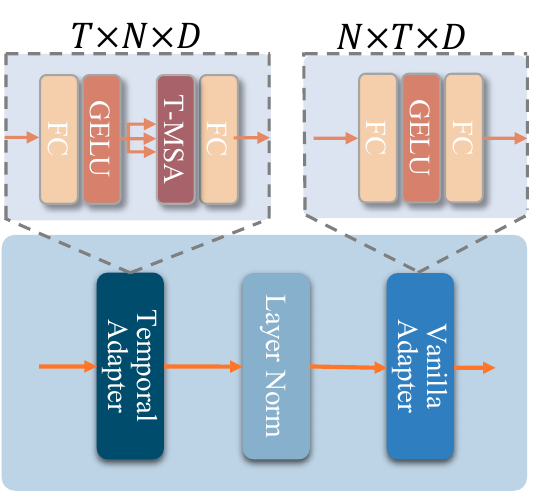}
		\put(2,  5){\small TMA}
		\put(4,  27){ \small $\bm{\mathcal{H}}^l $}
		\put(82,  27){ \small $\bm{\mathcal{T}}^{l+1} $}
	\end{overpic}
	\caption{\textbf{Temporal-Modeling Adapter (TMA) for temporal adaptation.} The input $\bm{\mathcal{H}}^l \in \mathbb{R}^{T \times N \times D}$ is fed into a Temporal Adapter to capture temporal information, then it is fed into a LayerNorm and a Vanilla Adapter to reduce the domain gap between SFER and DFER. $\bm{\mathcal{T}}^{l+1} \in \mathbb{R}^{T \times N \times D}$ is learned temporal information.}
	\label{fig:TMA}
\end{figure}

The model has developed a strong image-level representational ability through
training on the SFER dataset, but it can only model spatial dimension frame by
frame. To expand its ability for efficiently capturing the dynamic changes of
facial expressions in videos, we propose a Temporal-Modeling Adapter (TMA)
module. TMA is placed between transformer layers and comprises three
components: (1) a Temporal Adapter (T-Adapter), (2) a LayerNorm layer, and (3)
a Vanilla Adapter \cite{houlsby2019parameter}, as illustrated in Fig.
\ref{fig:TMA}.

The T-Adapter focuses on temporal modeling through the incorporation of
Temporal Multi-Head Self-Attention (T-MSA) \cite{Yang2023AIMAI}. Initially, the
video embeddings $\bm{\mathcal{H}}^l \in \mathbb{R}^{ T \times N \times D}$ is
reshaped to $\bm{\mathcal{H}}_T^l \in \mathbb{R}^{ N \times T \times D}$, then
it is downsampled to $\bm{\mathcal{H}}_T^{l'} \in \mathbb{R}^{ N \times T
		\times \gamma D}$ by a simple Linear layer and passed through a GELU activation
layer, where $\gamma$ is downsampling rate.
\begin{equation}
	\label{eq:download}
	\bm{\mathcal{H}}^{l'}_T =GELU(\bm{\mathcal{H}}^l_T\bm{W}^1_{d}+\bm{b}^1_{d}),
\end{equation}
where $\bm{W}^1_{d} \in \mathbb{R}^{D \times \gamma D}$, $ \bm{b}^1_d \in \mathbb{R}^{1 \times \gamma D}$ are learnable parameters of the downsampling Linear layer.
Next, $\bm{\mathcal{H}}_T^{l'}$ is fed into the T-MSA to capture the relationships between frames as follows:
\begin{equation}
	\bm{\mathcal{H}}_T^{l^{\prime \prime}}=T\text{-}M S A(\bm{\mathcal{H}}_T^{l^{\prime}}),
\end{equation}
whose processes are described by equations \ref{eq:SA}, \ref{eq:Head}, and \ref{eq:Cat}.
\begin{equation}
	S A(\boldsymbol{Q}, \boldsymbol{K}, \boldsymbol{V})=\operatorname{softmax}\left(\frac{\boldsymbol{Q} \boldsymbol{K}^{\top}}{\sqrt{\mathrm{D}_K}}\right) \boldsymbol{V},
	\label{eq:SA}
\end{equation}
\begin{equation}
	\operatorname{head}_i=S A\left(\boldsymbol{Q} \boldsymbol{W}_i^Q, \boldsymbol{K} \boldsymbol{W}_i^K, \boldsymbol{V} \boldsymbol{W}_i^V\right),
	\label{eq:Head}
\end{equation}
\begin{equation}
	\bm{\mathcal{H}}_T^{l^{\prime \prime}}= Cat \left(\operatorname{head}_1, \ldots, \text {head}_h\right) \boldsymbol{W}^{\boldsymbol{o}},
	\label{eq:Cat}
\end{equation}
where $\bm{Q}$, $\bm{K}$, $\bm{V}$ are the duplicates of $\bm{\mathcal{H}}^{l'}_T$ and the projections are parameter matrices:   $\{\bm{W}_i^Q,\bm{W}_i^K,\bm{W}_i^V \}\in \mathbb{R}^{D \times D_K}$, and $ \bm{W}^o \in \mathbb{R}^{h D_V \times D}$, $D_K=D_V=D / h$, $SA$ represents the Self-Attention mechanism, \( \text{head}_i \) corresponds to the \( i \)-th attention head, and $Cat$ denotes the concatenation operation. Finally, the $\bm{\mathcal{H}}_T^{l^{\prime \prime}}$ is upsampled to the original dimension as below:
\begin{equation}
	\bm{\mathcal{H}}^{l'''}_T = \bm{\mathcal{H}}^{l''}_T\bm{W}^1_{u}+\bm{b}^1_{u},
\end{equation}
where $\bm{W}^1_{u} \in \mathbb{R}^{\gamma D \times D}$, $ \bm{b}^1_d \in \mathbb{R}^{1 \times D}$ are learnable parameters of upsampling linear layer in T-Adapter.

After that, $\bm{\mathcal{H}}^{l'''}_T$ is reshaped back to
$\bm{\mathcal{H}}^{l''}$ with the shape of $T \times N \times D$ and fed into
LayerNorm Layer. To enhance the TMA's capacity to capture temporal information
and reduce the domain gap between SFER and DFER, we equip it with a Vanilla
Adapter \cite{houlsby2019parameter}. The Vanilla Adapter has two simple Linear
layers with a GELU activation function. The process can be described by the
following equations:
\begin{equation}
	\bm{\mathcal{H}}^{l''} = LayerNorm(\bm{\mathcal{H}}^{l''}),
\end{equation}
\begin{equation}
	\bm{\mathcal{T}}^{'} = GELU(\bm{\mathcal{H}}^{l''} \bm{W^2_d}+\bm{b^2_d}),
\end{equation}
\begin{equation}
	\bm{\mathcal{T}}^{l+1} = \bm{\mathcal{T}}^{'}\bm{W}^2_u+\bm{b}^2_u,
\end{equation}
where $\bm{W}^2_{d} \in \mathbb{R}^{D \times \gamma D}$, $\bm{W}^2_{u} \in \mathbb{R}^{\gamma D \times  D}$, $ \bm{b}^2_d \in \mathbb{R}^{1 \times \gamma D}$, $ \bm{b}^2_u \in \mathbb{R}^{1 \times D}$ are learnable parameters and $\bm{\mathcal{T}}^{l+1} \in \mathbb{R}^{T \times N \times D}$ is the temporal information tokens learned from $(l+1)$-th TMA.

With the integration of TMA modules, the image model can be efficiently
extended to a video model, with only 9M learnable parameters. As shown in Fig.
\ref{fig:network}, only the TMA, MCP, and Classifier are tunable, while all
other parameters are frozen during training.

\subsection{Emotion-Anchors based Self-Distillation Loss}
\label{section:loss}
\begin{figure}
	\centering
	\includegraphics[width=0.5\textwidth]{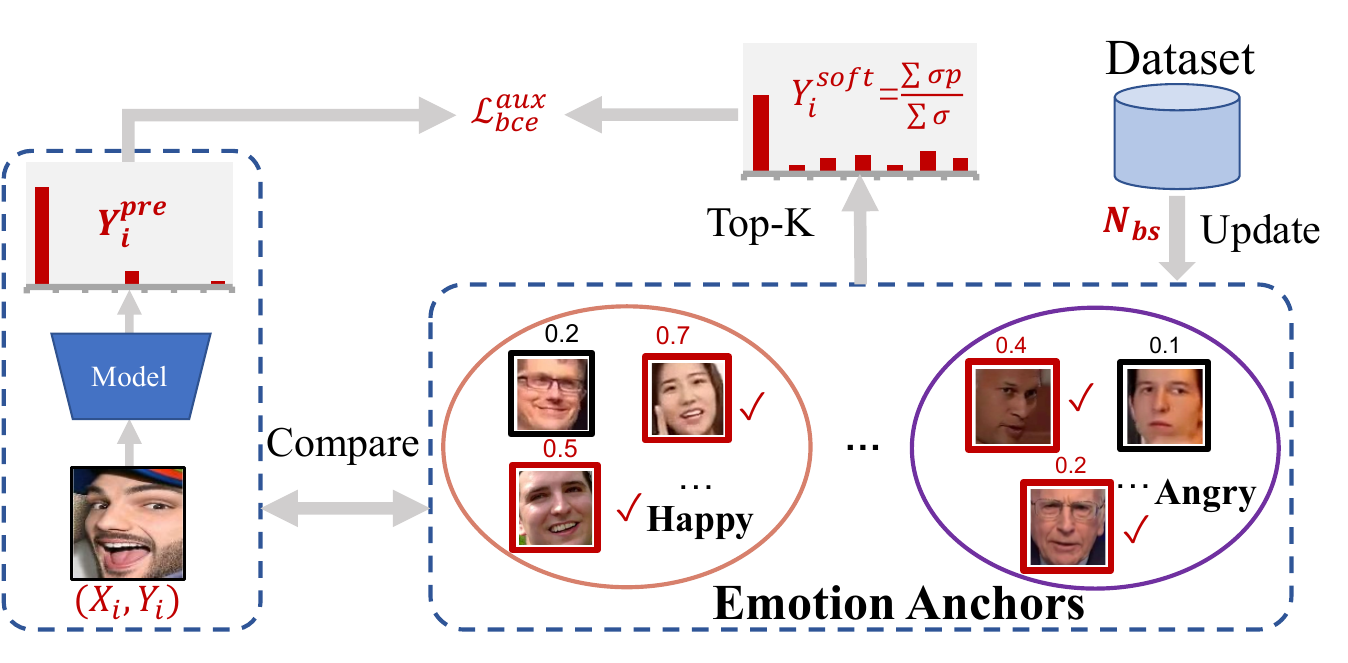}
	\caption{\textbf{Emotion-Anchors based Self-Distillation Loss.} $\bm{p}$ is output probabilities of S2D about emotion anchors. $\sigma$ is the similarity score between input and emotion anchors. ($\bm{X}_i$, $\bm{Y}_i$) is a sample from current batch and $\bm{Y}^{pre}_i$ is the corresponding predicted probability. $Y^{soft}_i $ is produced soft label for $\bm{X}_i$. }
	\label{fig:SDA}
\end{figure}

Due to the inherent complexity of facial expressions in videos and the subjectivity of annotators, manual annotations inevitably contain ambiguities. To mitigate the interference caused by the ambiguous annotations, we propose an Emotion-Anchors based Self-Distillation Loss as shown in Fig \ref{fig:SDA}. These reference samples for each emotion category in emotion anchors provide an auxiliary supervision signal to prevent ambiguous expression labels from deteriorating the performance of the FER model. We assume that the majority of the manual annotations are reliable and FER model can learn to recognize the correct samples gradually. Thus, we can select a bag of reference samples for each expression category to serve as emotion anchors. These anchors can be used to estimate a probability distribution of emotion and guide the model learning. Specifically, during the training process, we maintain two queues for each emotion, denoted as $\bm{Q}_c=\{\bm{v}^c_1,  \bm{v}_2^c,  \cdots,  \bm{v}_S^c\}$ and $\bm{P}_{c}=\{\bm{p}^c_1,  \bm{p}_2^c,  \cdots,  \bm{p}_S^c\}$, where $c \in \{1,  2,  \cdots,  C\}$,  $C$ is the number of emotion category,  and both $\bm{P}$ and $\bm{Q}$ have $S$ reference samples randomly selected from the training set.  Queue $\bm{Q}$ stores the model's output feature vectors $\bm{v}$ from the last transformer layer, while queue $\bm{P}$ stores the output probabilities $\bm{p}$ (normalized by softmax), where $\left\lVert \bm{p} \right\rVert = 1$. It's important to note that queues $\bm{P}$ and $\bm{Q}$ are not static, they are dynamically updated throughout the training process. Specifically, new reference samples are added to $\bm{P}$ and $\bm{Q}$ based on their ground truth labels from the current mini-batch, and the earlier samples are removed to maintain a fixed size. This dynamic updating mechanism ensures that the model is continually exposed to the most recent and relevant data, which helps in refining its understanding of each emotion category gradually. 

During training, for each sample $(\bm{X}_{i}, \bm{Y}_{i})$ in the current
mini-batch $\{(\bm{X}_{i}, \bm{Y}_{i})\}^{N_{bs}}_{i=1}$, the corresponding
output features and probabilities are denoted as $\bm{v_i}$ and $\bm{p_i}$,
respectively. As shown in Fig. \ref{fig:SDA}, we first calculate the cosine
similarity between $\bm{v_i}$ and each vector $\bm{v^c_j}$ in $\bm{Q}$,
yielding similarity scores between $\bm{v_i}$ and the samples in $C$ groups of
emotion anchors, $\bm{\alpha}=\{ \bm{\alpha}^1, \bm{\alpha}^2, \cdots,
	\bm{\alpha}^C \}$, where $\bm{\alpha}^c=\{\beta^c_{i, 1}, \beta^c_{i, 2},
	\cdots, \beta^c_{i, S}\}$. The similarity score $\beta^c_{i, j}$ between
$\bm{v}_i$ and $\bm{v}^c_j$ is computed using the formula:
\begin{equation}
	\beta^c_{i,  j}=\frac
	{\bm{v}^\top_{i} \cdot \bm{v}^c_{j}  }{\Vert \bm{v}_i  \Vert  \Vert \bm {v}^c_j\Vert }.
\end{equation}
Next,  we select the top $K$ samples in each emotion-anchor, resulting in the final scores $\bm{\sigma}=\{\sigma^1_{1},  \sigma^1_{2},  \cdots,  \sigma^{1}_K,    \cdots,  \sigma^C_1, \cdots, \sigma^C_K\}$, in which:
\begin{equation}
	\{\sigma^c_k\}_{k=1}^K = TopK(\bm{\alpha}^c, K).
\end{equation}
Afterwords, we obtain the soft label $\bm{Y}^{soft}_i$ for the current input sample $\bm{X}_i$:

\begin{equation}
	\bm{Y}^{soft}_i= \frac{1}{\sum^C_{c=1}\sum^K_{k=1}\sigma^c_{k}} \sum\limits_{c=1}^{C}\sum\limits_{k=1}^{K}\sigma^c_{k}  \bm{p}^{c}_{k}.
\end{equation}
Finally, we supervise the model using the labeled $\bm{Y}_i$ and the pseudo label $\bm{Y}^{soft}_i$, described as follows:
\begin{equation}
	\mathcal{L}^{target}_{ce}=\frac{1}{N} \sum^N_{i=1}CE(\bm{p}_i, \bm{Y}_i),
\end{equation}

\begin{equation}
	\mathcal{L}^{aux}_{bce}=\frac{1}{N} \sum^N_{i=1}BCE(\bm{p}_i, \bm{Y}^{soft}_i),
\end{equation}
where CE and BCE represent the cross-entropy loss and binary cross-entropy loss, respectively.
The total supervision loss is a combination of these two terms weighted by a hyperparameter $\eta  $:

\begin{equation}
	\mathcal{L} =\mathcal{L}^{target}_{ce}+\eta\mathcal{L}^{aux}_{bce}
	\label{eq:loss}.
\end{equation}

\section{Experiments}
\subsection{Datasets}
Our proposed model is evaluated on a diverse range of datasets: SFER datasets
include RAF-DB \cite{Li2017ReliableCA}, AffectNet-7/8
\cite{Mollahosseini2017AffectNetAD}, and FERPlus \cite{Barsoum2016TrainingDN},
and DFER datasets encompass DFEW \cite{jiang2020dfew}, FERV39K
\cite{wang2022ferv39K}, and MAFW \cite{Liu2022MAFWAL}. RAF-DB contains 29,672
real-world facial images annotated with basic expressions by 40 trained
annotators, using 12,271 for training and 3,068 for testing. AffectNet-7/8
comprises over 1 million images, manually annotating around 450,000.
AffectNet-7 has 7 basic expressions with 283,901 training and 3,500 testing
images, and AffectNet-8 adds the \emph{contempt} category with 287,568 training
and 4,000 testing images. FERPlus extends FER2013 with 28,709 training, 3,589
validation, and 3,589 testing images. DFEW involves 16,000 video clips
categorized into 7 expressions, using 12,059 clips for experimentation through
5-fold cross-validation. FERV39K has 38,935 video sequences from 4 scenarios,
31,088 clips for training, and 7,847 clips for testing. MAFW contains 10,045
video clips annotated with 11 emotions, focusing on the video modality with
9,172 clips for experiments, following a 5-fold cross-validation protocol.

\subsection{Evaluation Metrics}
We evaluate the performance on SFER datasets with Accuracy, which has been
widely used for SFER tasks. As for DFER datasets, we follow the approach of
previous studies
\cite{Zhao2021FormerDFERDF,li2023intensity,wang2023rethinking,sun2023mae} and
report both unweighted average recall (UAR), which represents the averaged
accuracy across classes, and weighted average recall (WAR), which indicates the
overall accuracy (the same as the metric of SFER).
\subsection{Implementation Details}
In all experiments, we extract facial landmark-aware features with
MobileFaceNet \cite{Chen2018MobileFaceNetsEC}, a widely used facial landmark
detection tool, by default. ViT-B/16 \cite{Dosovitskiy2020AnII} is used as the
backbone of S2D. We follow the same training strategy as Li et al.
\cite{li2023} to train our SFER model. For the DFER model, we initialized it
with the weights pre-trained on the AffectNet-7 dataset and fine-tuned it on
the DFER dataset. Following \cite{li2023intensity, Zhao2021FormerDFERDF,
	Ma2022SpatioTemporalTF, Liu2022MAFWAL}, we sample 16 frames from each video
clip and resize them to 224$\times$224. Incidentally, we initially perform crop
and alignment on the MAFW dataset using the RetinaFace
\cite{deng2020retinaface} model from the deepface toolkit \footnote{deepface
	toolkit: \url{https://github.com/serengil/deepface}}
\cite{serengil2020lightface}. Regarding training strategy, we mainly follow
MAE-DFER \cite{sun2023mae}. Specifically, we employ an AdamW optimizer with
$\beta_1=0.9$ and $\beta_2=0.95$, an overall batch size of 64, a base learning
rate $\operatorname{lr}_{base}$ of $1e - 5$, and a weight decay of 0.05. We
linearly scale the base learning rate according to the overall batch size $N
		_{bs}$, using the formula: $\operatorname{lr}= \operatorname{lr}_{base} \times
	\frac{N_{bs}}{8}$. Furthermore, we use cosine annealing to decay the learning
rate with the model undergoing training for a total of 100 epochs. The
downsampling rate $\gamma$ in Eq. \ref{eq:download} is set to 0.25 and the
hyperparameter $\eta $ in Eq. \ref{eq:loss} is gradually increased from 0 to 1.
For all DFER datasets, the queue size $S$ for each emotion category is set to
16, and the $K$ is set to 2. During training, only the parameters of MCPs,
TMAs, and the Classifier are tunable, while all other parameters are frozen.
All experiments are conducted on 8 Nvidia 3090Ti GPUs using the PyTorch
framework.

\subsection{Ablation Studies}
In this section, we conduct ablation experiments on AffectNet-7/8
\cite{Mollahosseini2017AffectNetAD}, RAF-DB \cite{Li2017ReliableCA}, FERPlus
\cite{Barsoum2016TrainingDN}, DFEW \cite{jiang2020dfew} and FERV39K
\cite{wang2022ferv39K} datasets to demonstrate the effectiveness of several key
components in S2D. For simplicity, we only report the results
	of fold 1 (fd1) for DFEW, following the practice of prior work
	\cite{sun2023mae}. 

\subsubsection{Image-Level Representation Enhancement}

\begin{table}[]
	\caption{Comparison of cross-dataset evaluation. AffectNet-Small is sampled from AffectNet-7 and has the same data size as RAF-DB. Underlined results are produced by the model initialized with MAE-ImageNet1K \cite{he2022masked}, while bolded results are from the model initialized with MAE-AffectNet.  All results are produced by SFER models (the S2D without TMA and SDL).}
	\resizebox{\linewidth}{!}{
		\begin{tabular}{cccccccc}
			\toprule
			\multirow{2}{*}{\tabincell{c}{Training \\ images}} & \multirow{2}{*}{\tabincell{c}{Pre-training \\data}} & \multicolumn{6}{c}{Testing dataset}  \\  \cmidrule(lr) { 3-8}
			        &                   & \multicolumn{2}{c}{RAF-DB} & \multicolumn{2}{c}{AfectNet-7}      & \multicolumn{2}{c}{FERPlus}                                                                                              \\
			\midrule
			12,271  & RAF-DB            & \textbf{92.21}             & \underline{85.07} & \textbf{47.57}              & \underline{46.83} & \textbf{77.35} & \underline{62.82} \\
			283,901 & AffectNet-7       & \textbf{76.16}             & \underline{70.73} & \textbf{66.42}              & \underline{63.67} & \textbf{64.32} & \underline{55.89} \\
			12,271  & AffectNet-7-Small & \textbf{69.13}             & \underline{60.40} & \textbf{61.40}              & \underline{53.44} & \textbf{62.08} & \underline{48.21} \\
			\bottomrule
		\end{tabular}
	}
	\label{tab:cross_dataset}
\end{table}

\textbf{\\Selection of static facial expression features.}
Cross-dataset evaluations were initially conducted on the RAF-DB and AffectNet-7 datasets to assess their respective robustness.
Table \ref{tab:cross_dataset} displays our SFER models' cross-dataset evaluation results (highlighted in bold), which revealed a significant performance decline of the SFER model pre-trained on RAF-DB when tested on the AffectNet-7 dataset, culminating in a 44.64\%  drop in accuracy. Conversely, the model pre-trained on AffectNet-7, when evaluated on the RAF-DB dataset, demonstrated an accuracy improvement of 9.74\%. It is well known that the AffectNet dataset has a higher noise level than the RAF-DB dataset. To explore whether this difference arises from data size or sample diversity, a further experiment about AffectNet was undertaken using an equivalent amount of data as RAF-DB. The subset sampled from AffectNet-7 is named AffectNet-Small. Despite AffectNet-Small having a much smaller data size than AffectNet-7, it only experienced a 5.02\% accuracy drop on AffectNet-7 and still performs well on RAF-DB. For objectivity and fairness, a third-party evaluation was also introduced using the FERPlus testing set, consisting of 3,589 gray-scale images, each sized to 48x48 pixels. As shown in Table \ref{tab:cross_dataset}, testing on FERPlus reveals a pronounced 14.86\% accuracy drop for the SFER model pre-trained on RAF-DB, but only a slight drop of 2.1\% accuracy for the model pre-trained on AffectNet-7. Interestingly, the model trained on RAF-DB outperforms AffectNet-7 when tested on the FERPlus dataset. We hypothesize that this performance disparity is attributable to the greater consistency in annotation methods between RAF-DB and FERPlus, compared to AffectNet. To eliminate the influence of initial weights, we conducted additional experiments using MAE-ImageNet1K \cite{he2022masked} weights for initialization. The experimental results, as highlighted in Table \ref{tab:cross_dataset}, are consistent with our previous observations. Specifically, the model trained on AffectNet-7 shows a lesser performance drop when tested on other datasets compared to the model trained on RAF-DB, indicating that the model trained on AffectNet-7 seems to be more robust than RAF-DB.

Subsequent experiments were also conducted on DFER datasets by initializing the
DFER model using various SFER models. These SFER models were
	initialized with the weights pre-trained on
	ImageNet-1K\cite{russakovsky2015imagenet} or AffectNet with
	MAE\cite{he2022masked} pre-training method and fine-tuned on RAF-DB, FERPlus,
	and AffectNet-7, respectively. This was done to evaluate the generalization
ability of the static appearance features obtained through pre-training on
various SFER datasets. As shown in Table
	\ref{tab:mcp_different_sfer_dataset}, the DFER model pre-trained on the
	AffectNet-7 dataset exhibits superior performance on both DFEW and FERV39K
	datasets, particularly when the SFER model is initialized with general domain
	weights. Given that the model pre-trained on AffectNet-7 outperforms others on
	DFEW and FERV39K, and considering the larger data size and greater diversity of
	samples in the AffectNet-7 dataset, we recommend utilizing the SFER model
	pre-trained on AffectNet-7 for enhanced downstream generalization performance. 
\begin{table}[]
	\centering
	\caption{Comparison of different pre-training SFER data. All results are produced by S2D with SDL loss.  As AffectNet-7 has a larger data size and more diverse samples, we recommend using the SFER model pre-trained on AffectNet-7 dataset for better generalization performance.
	}

	\begin{tabular}{cccccc}
		\toprule
		\multirow{2}{*}{Init weights} & \multirow{2}{*}{\tabincell{c}{Pre-training                                                                                                         \\data }} & \multicolumn{2}{c}{DFEW} & \multicolumn{2}{c}{FERV39K} \\ \cmidrule(lr) { 3 - 4 }\cmidrule(lr) { 5-6}
		                              &                                            & WAR                     & UAR                     & UAR                     & WAR                     \\
		\midrule
		\multirow{3}{*}{MAE-ImageNet1K \cite{he2022masked}}
		                              & RAF-DB                  & 55.41 & 69.50 & 34.25 & 46.77 \\
		                              & FERPlus                 & 56.10 & 70.10 & 33.88 & 46.71 \\
		                              & AffectNet-7              & 56.69 & 70.91 & 34.83 & 47.36 \\
		\midrule
		\multirow{3}{*}{MAE-AffectNet \cite{he2022masked}}
		                              & RAF-DB                                     & 61.06                   & 75.22                   & 38.97                   & 51.50                   \\
		                              & FERPlus                                    & 61.62                   & 76.08                   & 39.15                   & 51.56                   \\
		                              & AffectNet-7                                & 61.56                   & 76.16                   & 41.28                   & 52.56                   \\
		\bottomrule
	\end{tabular}
	\label{tab:mcp_different_sfer_dataset}
\end{table}

\begin{table}
	\centering
	\caption{Ablation on different facial landmark detection models with varying levels of localization precision. ION (Inter-Ocular Normalization) is an evaluation metric for landmark detection. The results are produced by S2D with SDL.}
	\resizebox{\linewidth}{!}{\begin{tabular}{ccccccc}
			\toprule
			\multirow{2}{*}{ Model }                                           & \multirow{2}{*}{ ION $\downarrow$ } & \multirow{2}{*}{\tabincell{c}{Params                                                                                                         \\(M)} }& \multicolumn{2}{c}{ DFEW } & \multicolumn{2}{c}{ FERV39k }       \\ \cmidrule(lr) { 4-5 }\cmidrule(lr) { 6-7}
			                                                                   &                                     &                                      & UAR                     & WAR                     & UAR                     & WAR                     \\
			\midrule
			 MobileNetV2@56 \cite{Sandler2018MobileNetV2IR}  & 5.29              & 3.74               & 60.74 & 75.31 & 39.13 & 51.28 \\
			 MobileNetV2@224 \cite{Sandler2018MobileNetV2IR} & 4.39              & 3.74               & 61.45 & 75.69 & 39.76 & 51.35 \\
			PFLD \cite{Guo2019PFLDAP}                                          & 3.97                                & 0.73                                 & 61.29                   & 76.33                   & 39.33                   & 51.55                   \\
			MobileNetV2\_ED \cite{Sandler2018MobileNetV2IR}                    & 3.96                                & 3.74                                 & 62.05                   & 76.25                   & 39.38                   & 51.37                   \\
			MobileFaceNet \cite{Chen2018MobileFaceNetsEC}                      & 3.76                                & 1.01                                 & 61.56                   & 76.16                   & 41.28                   & 52.56                   \\
			\bottomrule
		\end{tabular}
	}
	\label{tab:different_landmark}
\end{table}

\textbf{\\Ablation study on the landmark-aware features.}
We compare the performance of three lightweight facial landmark detection models from PyTorch Face Landmark toolkit\footnote{PyTorch Face Landmark: \url{https://github.com/cunjian/pytorch_face_landmark}} on extracting landmark-aware features for in-the-wild DFER tasks. These models are PFLD \cite{Guo2019PFLDAP}, MobileNetV2@56 \cite{Sandler2018MobileNetV2IR}, MobileNetV2@224 \cite{Sandler2018MobileNetV2IR}, MobileNetV2\_ED \cite{Sandler2018MobileNetV2IR}, and MobileFaceNet \cite{Chen2018MobileFaceNetsEC}, which have different localization precision and complexity levels. Table \ref{tab:different_landmark} shows that the features extracted by these models perform well on both DFEW and FERV39K datasets.  However, as the localization precision of the facial landmark detection model decreases, the performance of the model declines to varying degrees on both DFEW and FERV39K, with the decline being particularly noticeable on DFEW. Considering that the MobileFaceNet  \cite{Chen2018MobileFaceNetsEC} model has the lowest inter-ocular normalization error (ION) and achieves the highest performance on the more challenging and representative FERV39K dataset, we choose it as the facial landmark model for extracting landmark-aware features in the following experiments.

\begin{table}
	\centering
	\caption{Ablation study on the Multi-View Complementary Prompter. The results on SFER datasets are produced by our SFER models (the S2D without TMA and SDL), while the results on DFER datasets are produced by our S2D with SDL. None: without facial landmark-aware features. MCP: fuse static features and landmark-aware features with the Multi-View Complementary prompter. CAP: fuse static features and facial landmark-aware features with the commonly used concatenation+projection module.  }

	\resizebox{0.9\linewidth}{!}{
		\begin{tabular}{cccccc}
			\toprule \multirow{5}{*}{ SFER } & Fusion Method                    & RAF-DB                     & AffectNet-7                   & \multicolumn{2}{c}{ FERPlus }         \\\cmidrule(lr){ 2 - 6 }
			                                 & None                             & 91.07                      & 66.09                         & \multicolumn{2}{c}{90.18}             \\
			                                 & MCP                              & 92.21                      & 66.42                         & \multicolumn{2}{c}{91.01}             \\
			                                 & CAP                              & 88.88                      & 64.22                         & \multicolumn{2}{c}{86.15}             \\
			\midrule \multirow{7}{*}{ DFER } & \multirow{3}{*}{ Fusion Method } & \multicolumn{2}{c}{ DFEW } & \multicolumn{2}{c}{ FERV39K }                                         \\\cmidrule(lr) { 3 - 4 }\cmidrule(lr) { 5 - 6 }
			                                 &                                  & UAR                        & WAR                           & UAR                           & WAR   \\\cmidrule(lr) { 2 - 6 }
			                                 & None                             & 61.06                      & 75.48                         & 38.74                         & 51.44 \\
			                                 & MCP                              & 61.56                      & 76.16                         & 41.28                         & 52.56 \\
			                                 & CAP                              & 59.94                      & 73.56                         & 38.31                         & 50.76 \\
			\bottomrule
		\end{tabular}
	}
	\label{tab:mcp_landmark}

\end{table}

\subsubsection{Ablation study on the Multi-View Complementary Prompter}
As presented in Table \ref{tab:mcp_landmark}, we evaluated the impact of
different fusion methods on SFER and DFER testing sets. Compared to the
baseline (line 1), incorporating facial landmark-aware features via MCP yielded
notable improvements (+1.14\% on RAF-DB, +0.33\% on AffectNet-7, +0.83\% on
FERPlus, +0.68\% WAR on DFEW and +1.12\% WAR on FERV39K) in performance across
both SFER and DFER datasets. This emphasizes the significance of facial
landmark-aware features in both SFER and DFER tasks. Furthermore, replacing MCP
with the common fusion module of concatenation and projection (CAP, line 3)
resulted in marked performance declines across all datasets. This suggests that
MCP is more effective for integrating static features and landmark-aware
features, whereas CAP may potentially destroy the structure of the original
static features and degrade model performance.

\begin{table}
	\centering
	\caption{Ablation study on the different adapters. The proposed Temporal-Modeling Adapter (TMA) achieves the best results.  }
	\begin{tabular}{ccccc}
		\toprule
		\multirow{2}{*}{ Adapter }                  & \multicolumn{2}{c}{ DFEW } & \multicolumn{2}{c}{ FERV39k }                 \\ \cmidrule(lr) { 2 - 3 }\cmidrule(lr) { 4-5 }
		                                            & UAR                        & WAR                           & UAR   & WAR   \\
		\midrule
		None                                        & 54.00                      & 68.60                         & 33.17 & 46.03 \\
		Vanilla Adapter \cite{houlsby2019parameter} & 58.67                      & 72.40                         & 38.61 & 50.91 \\
		Temporal Adapter                            & 61.32                      & 75.99                         & 39.00 & 51.15 \\
		Temporal-Modeling Adapter                   & 61.56                      & 76.16                         & 41.28 & 52.56 \\
		\bottomrule
	\end{tabular}
	\label{tab:ablation_adapter}
\end{table}

\subsubsection{Ablation study on the Temporal-Modeling Adapter}
We evaluate the model's performance with various adapters and report the
results in Table \ref{tab:ablation_adapter}. The results reveal that the
Vanilla Adapter \cite{houlsby2019parameter} brings 4.67\% UAR with 3.80\% WAR
and 5.44\% UAR with 4.88\% WAR improvements on DFEW and FERV39K testing sets,
respectively. Although this signifies a positive impact on the model's
performance, it remains somewhat limited due to its lack of temporal
information modeling. In contrast, the Temporal Adapter exhibits significant
improvements over the Vanilla Adapter, surpassing it by 2.65\% UAR with 3.59\%
WAR on DFEW, and 0.39\% UAR with 0.24\% WAR on the FERV39K dataset. \emph{This
enhancement can be attributed to the robust temporal modeling capabilities
brought by T-MSA in the Temporal Adapter.} Consequently, we combine the Vanilla
Adapter with the Temporal Adapter, and introduce the Temporal-Modeling Adapter
(TMA), leading to further performance improvements, particularly on FERV39K
(+2.28\% UAR and +1.41\% WAR).

\begin{table}
	\centering
	\caption{The effectiveness of SDL. We compare SDL with the original One-Hot supervised signal and the commonly used Label Smoothing method. }
	\begin{tabular}{ccccc}
		\toprule
		\multirow{2}{*}{ Method } & \multicolumn{2}{c}{ DFEW } & \multicolumn{2}{c}{ FERV39k }                 \\ \cmidrule(lr) { 2 - 3 }\cmidrule(lr) { 4-5 }
		                          & UAR                        & WAR                           & UAR   & WAR   \\
		\midrule
		One-Hot                   & 61.50                      & 76.12                         & 41.48 & 52.13 \\
		Label Smoothing           & 61.55                      & 75.99                         & 40.68 & 51.93 \\
		One-Hot+SDL (ours)        & 61.56                      & 76.16                         & 41.28 & 52.56 \\
		\bottomrule
	\end{tabular}
	\label{tab:label_distribution}
\end{table}

\subsubsection{The Effectiveness of SDL}
To validate the effectiveness of the auxiliary supervised signal provided by
our proposed SDL in reducing the interference caused by ambiguous annotations,
we compare it with the original One-Hot supervision signal and the commonly
used Label Smoothing method. As Table \ref{tab:label_distribution} shows, Label
Smoothing slightly decreases the performance of WAR ($-0.13$\% WAR on DFEW and
$-0.20$\% WAR on FERV39K). In contrast, our SDL considerably improves the WAR
on DFEW and FERV39K by 0.04\% and 0.43\%, respectively. We further visualize
the output probability distribution of the model with SDL in Fig.
\ref{fig:prob}, which demonstrates that SDL significantly enhances the model's
discriminability for ambiguous samples. The observed decline in UAR on FERV39K
is attributable to data imbalance. Based on experimental experience, the model
exhibits a stronger emphasis on learning expressions from specific classes with
larger data volumes. While WAR is sensitive to the accuracy of these
categories, UAR is primarily influenced by the minor categories. Therefore, an
increase in WAR often comes at the expense of a decrease in UAR.

\begin{table}[!t]
	\caption{The contribution of our proposed components. MCP: multi-view complementary prompter; TMA: temporal-modeling adapter; SDL: emotion-anchors based self-distillation loss.  }
	\begin{tabular}{ccccccc}
		\toprule
		\multirow{2}{*}{ MCP } & \multirow{2}{*}{ TMA } & \multirow{2}{*}{ SDL } & \multicolumn{2}{c}{ DFEW } & \multicolumn{2}{c}{ FERV39k }                 \\ \cmidrule(lr) { 4-5 } \cmidrule(lr) { 6 - 7 }
		                       &                        &                        & UAR                        & WAR                           & UAR   & WAR   \\
		\midrule
		                       &                        &                        & 51.05                      & 65.78                         & 32.12 & 44.81 \\
		\checkmark             &                        &                        & 54.27                      & 68.73                         & 33.33 & 46.03 \\
		\checkmark             & \checkmark             &                        & 61.50                      & 76.12                         & 41.48 & 52.13 \\
		\checkmark             & \checkmark             & \checkmark             & 61.56                      & 76.16                         & 41.28 & 52.56 \\
		\toprule
	\end{tabular}
	\label{tab:ablation_dfer}
\end{table}

\begin{figure}[!t]
	\centering
	\includegraphics[width=0.5\textwidth]{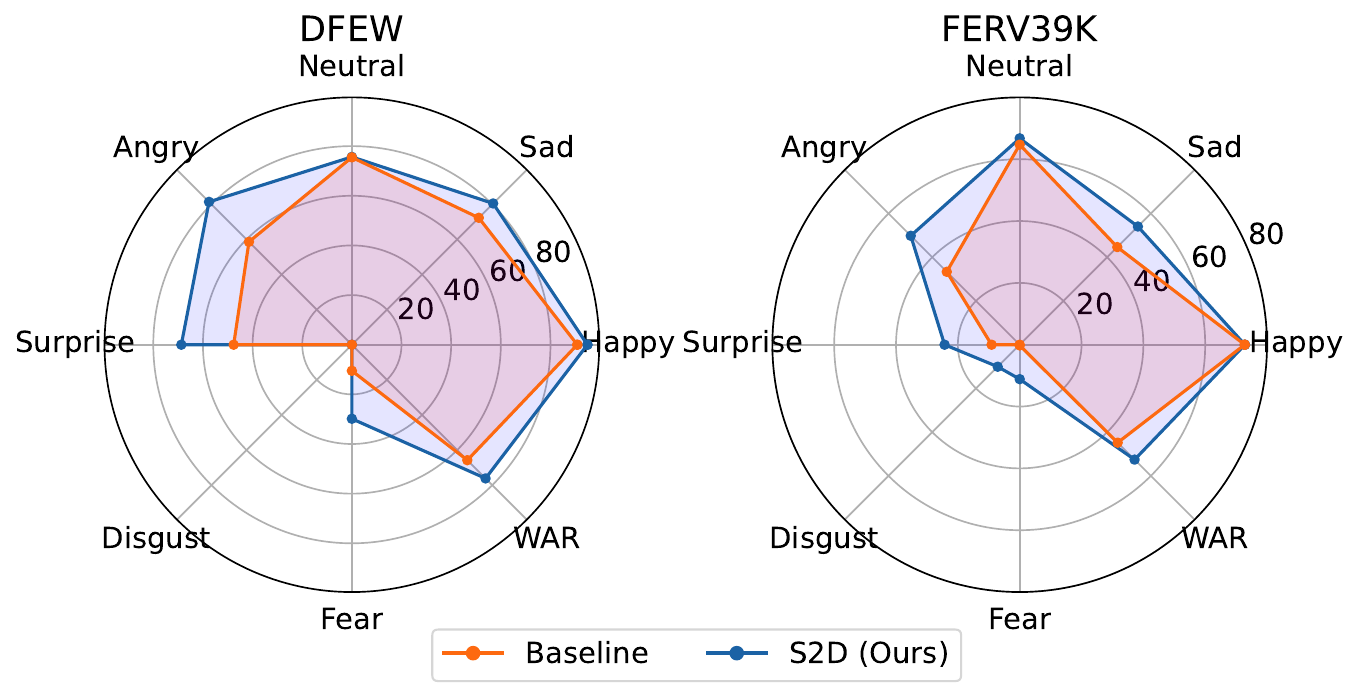}
	\caption{\textbf{The comparison of our proposed model with baseline at class level.} We visualize the overall accuracy (WAR) and class accuracy of each emotion on DFEW (fd1) and FERV39K datasets. The baseline method is our model without TMA, MCP, and SDL. (i.e., line 1 in Table \ref{tab:ablation_dfer}).}
	\label{fig:gender}
\end{figure}

\begin{table}[!t]
	\centering
	\caption{Compare with state-of-the-art SFER methods on RAF-DB,  AffectNet-7, AffectNet-8 and FERPlus testing sets. We highlight the best result in bold and underline the second-best result. $\ddagger$ means S2D pre-trained on the AffectNet-8 dataset. The results are produced by the SFER model (the S2D model without TMA and SDL). Our method achieved comparable or even better performance than previous state-of-the-art SFER methods.}
	\begin{center}
		\resizebox{\linewidth}{!}{
			\begin{tabular}{lcccc}
				\toprule
				Method                                    & RAF-DB            & AffectNet-7       & AffectNet-8       & FERPlus           \\
				\midrule
				SCN \cite{wang2020suppressing}            & 88.14             & -                 & 60.23             & 89.35             \\
				EAC \cite{zhang2022learn}                 & 88.99             & 65.32             & -                 & 89.64             \\
				DMUE \cite{she2021dive}                   & 88.42             & -                 & 63.11             & 89.51             \\
				LDLVA \cite{le2023uncertainty}            & 90.51             & 66.23             & -                 & -                 \\
				MVT \cite{li2021mvt}                      & 88.62             & 64.57             & 61.40             & 89.22             \\
				VTFF \cite{ma2021facial}                  & 88.14             & -                 & 61.58             & 88.81             \\
				TransFER \cite{Xue2021TransFERLR}
				                                          & 90.91             & 66.23             & -                 & 90.83             \\
				CLIPER \cite{Li2023CLIPERAU}              & 91.61             & 66.29             & 61.98             & -                 \\
				APViT \cite{xue2022vision}                & 91.98             & 66.91             & -                 & 90.86             \\
				TAN	\cite{ma2023transformer}              & 90.87             & 66.45             & -                 & 91.00             \\
				PF-ViT \cite{li2023}                      & 92.07             & \underline{67.23} & \textbf{64.10}    & \underline{91.16} \\
				\midrule
				\textbf{S2D w/o TMA (ours)}               & \underline{92.21} & 66.42             & \underline{63.76} & 91.01             \\
				\textbf{S2D w/o TMA (ours)  $^\ddagger $} & \textbf{92.57}    & \textbf{67.62}    & 63.06             & \textbf{91.17}    \\
				\bottomrule
			\end{tabular}
		}
	\end{center}
	\label{tab:result_sfer}
\end{table}

\subsubsection{Contribution of Proposed Components}
In Table \ref{tab:ablation_dfer}, we conducted ablation experiments on the DFEW
and FERV39K testing sets to quantify the contributions of the proposed
components. Compared to the baseline (line 1), incorporating the MCP (line 2)
improved the performance by 3.22\% UAR and 2.95\% WAR on DFEW, and 1.21\% UAR
and 1.22\% WAR on FERV39K. This indicates the efficacy of facial landmark-aware
features in enhancing the model's image-level representation capabilities.
Moreover, integrating the Temporal-Modeling Adapter (TMA) module (line 3)
further enhanced performance (+7.23\% UAR and +7.39\% WAR on DFEW, +9.36\% UAR
and +6.10\% WAR on FERV39K), attributed to its powerful ability to capture the
temporal information of facial expressions. The SDL (line 4) still improved the
performance by 0.04\% WAR on DFEW and 0.43\% WAR on FERV39K, even after S2D
achieved significant improvement over the baseline. This highlights the
significance of SDL, particularly in mitigating the interference caused by
ambiguous annotations. Overall, the results in Table \ref{tab:ablation_dfer}
demonstrate the effectiveness of each component and their collective
contribution to S2D. Furthermore, to specifically highlight
the comparative advantage of our proposed method over the baseline, we have
visualized the overall accuracy (WAR) and the detailed breakdown of class
accuracy for each emotion category on the DFEW and FERV39K datasets in Fig.
\ref{fig:gender}. This visualization clearly demonstrates the method's enhanced
efficacy at the class level when compared directly with the baseline. %

\subsection{Comparison with State-of-the-Art Methods}
We compared our method with existing state-of-the-art methods on two sets of
datasets: SFER datasets (RAF-DB \cite{Li2017ReliableCA}, AffectNet
\cite{Mollahosseini2017AffectNetAD}, and FERPlus \cite{Barsoum2016TrainingDN})
and DFER datasets (DFEW \cite{jiang2020dfew}, FERV39K \cite{wang2022ferv39K},
and MAFW \cite{Liu2022MAFWAL}). In this section, we report the
	average results across all 5 folds for DFEW and MAFW datasets. 
\subsubsection{Results on SFER Datasets}
Table \ref{tab:result_sfer} provides a comparison with several SOTA methods on
popular SFER datasets. Our method achieved the best results on RAF-DB,
AffectNet-7, and FERPlus. Particularly noteworthy is our achievement of 92.57\%
accuracy on RAF-DB, surpassing existing methods by 0.50\%. Despite the limited
quantity of data in FERPlus and the small gray image size of 48x48 (meaning an
obvious domain gap), our method still achieved the highest accuracy.

\begin{table*}[t]
	\centering
	\caption{Comparison with state-of-the-art methods on DFEW,  FERV39K and MAFW.  *: sampling two clips uniformly along the temporal axis for each video sample and then calculating the average score as the final prediction during testing time. ${ }^{\dagger}$: training with oversampling strategy. UAR: unweighted average recall; WAR: weighted average recall. }
	\begin{tabular}{lcccccccc}
		\toprule
		\multirow{2}{*}{ Method }                         & \multirow{2}{*}{\tabincell{c}{Tunable                                                                                                                         \\ Params (M)}} & \multicolumn{2}{c}{ DFEW }        & \multicolumn{2}{c}{ FERV39k } &       \multicolumn{2}{c}{ MAFW }                \\ \cmidrule(lr) { 3-4 } \cmidrule(lr) { 5 - 6 }\cmidrule(lr) { 7-8 }
		                                                  &                                       & UAR               & WAR               & UAR               & WAR               & UAR               & WAR               \\
		\midrule
		\textit{\textbf{Supervised learning models}}      &                                       &                   &                   &                   &                   &                   &                   \\
		C3D \cite{Tran2014LearningSF}                     & 78                                    & 42.74             & 53.54             & 22.68             & 31.69             & 31.17             & 42.25             \\
		R(2+1)D-18  \cite{tran2018closer}                 & 33                                    & 42.79             & 53.22             & 31.55             & 41.28             & -                 & -                 \\
		3D ResNet-18 \cite{He2015DeepRL}                  & 33                                    & 46.52             & 58.27             & 26.67             & 37.57             & -                 & -                 \\
		ResNet-18+LSTM  \cite{Zhao2021FormerDFERDF}       & -                                     & 51.32             & 63.85             & 30.92             & 42.59             & 28.08             & 39.38             \\
		Former-DFER \cite{Zhao2021FormerDFERDF} [MM'21]   & 18                                    & 53.69             & 65.70             & 37.20             & 46.85             & -                 & -                 \\
		CEFLNet  \cite{liu2022clip}  [IS'2022]            & 13                                    & 51.14             & 65.35             & -                 & -                 & -                 & -                 \\
		T-ESFL \cite{Liu2022MAFWAL}  [MM'22]              & -                                     & -                 & -                 & -                 & -                 & 33.28             & 48.18             \\
		EST \cite{liu2023expression} [PR'23]              & 43                                    & 53.43             & 65.85             & -                 &                   & -                 & -                 \\
		STT \cite{Ma2022SpatioTemporalTF}  [arXiv'22]     & -                                     & 54.58             & 66.65             & 37.76             & 48.11             & -                 & -                 \\
		NR-DFERNet \cite{li2022nr} [arXiv'22]             & -                                     & 54.21             & 68.19             & 33.99             & 45.97             & -                 & -                 \\
		IAL \cite{li2023intensity} [AAAI'23]              & 19                                    & 55.71             & 69.24             & 35.82             & 48.54             & -                 & -                 \\
		M3DFEL \cite{wang2023rethinking} [CVPR'23]        & -                                     & 56.10             & 69.25             & 35.94             & 47.67             & -                 & -                 \\
		\textbf{S2D (ours)}                               & 9                                     & 62.57             & \underline{75.98} & 40.89             & 51.83             & 39.87             & \underline{56.20} \\
		\textbf{S2D (ours)  * }                           & 9                                     & 61.82             & \textbf{76.03}    & 41.28             & \textbf{52.56}    & \underline{41.86} & \textbf{57.37}    \\
		\textbf{S2D (ours)  *${ }^{\dagger}$ }            & 9                                     & \textbf{65.45}    & 74.81             & \textbf{43.97}    & 46.21             & \textbf{43.40}    & 55.22             \\
		\midrule
		\textit{\textbf{Self-supervised learning models}} &                                       &                   &                   &                   &                   &                   &                   \\
		MAE-DFER \cite{sun2023mae} *  [MM'23]             & 85                                    & \underline{63.41} & 74.43             & \underline{43.12} & \underline{52.07} & 41.62             & 54.31             \\
		\midrule
		\textit{\textbf{Vision-language models}}          &                                       &                   &                   &                   &                   &                   &                   \\
		CLIPER \cite{Li2023CLIPERAU} [arXiv'23]           & 88                                    & 57.56             & 70.84             & 41.23             & 51.34             & -                 & -                 \\
		DFER-CLIP \cite{zhao2023prompting} [BMVC'23]      & 90                                    & 59.61             & 71.25             & 41.27             & 51.65             & 38.89             & 52.55             \\
		\bottomrule
		\\
	\end{tabular}

	\label{tab:result_dfer}
\end{table*}

\subsubsection{Results on DFER Datasets}
Firstly, we compare S2D with previous state-of-the-art supervised learning
models on DFEW, FERV39K, and MAFW. Table \ref{tab:result_dfer} shows that S2D
significantly outperformed the previous best methods (i.e., IAL
\cite{li2023intensity}, M3DFEL \cite{wang2023rethinking}) and achieved a
notable improvement of 6.47\% UAR with 6.73\% WAR and 5.05\% UAR with 4.16\%
WAR on DFEW and FERV39K datasets, respectively. Regarding MAFW, S2D achieves a
new state-of-the-art performance, surpassing the previous best methods
\cite{Liu2022MAFWAL} by a significant margin of 6.59\% in UAR and 9.19\% in
WAR. These significant improvements indicate that our method can learn powerful
representations for DFER by pre-training on a large-scale SFER dataset.

Secondly, we compare S2D with the self-supervised method MAE-DFER
\cite{sun2023mae}, which was pre-trained on a large-scale video dataset (over 1
million video clips) with a self-supervised approach \cite{tong2022videomae}.
To ensure fairness, we adopted the same testing strategy as MAE-DFER.
Experimental results demonstrate that our method outperformed MAE-DFER by
1.60\%, 0.49\%, and 3.06\% in WAR on DFEW, FERV39K, and MAFW respectively, with
fewer tunable parameters and training costs (refer to Fig. \ref{fig:params}).
Furthermore, our method, trained with an oversampling strategy, also
outperformed MAE-DFER in UAR, achieving improvements of 2.04\% on DFEW, 0.85\%
on FERV39K, and 1.54\% on MAFW.

We also compare S2D with two vision-language models CLIPER
\cite{Li2023CLIPERAU} and DFER-CLIP \cite{zhao2023prompting}, which rely on
both prior language knowledge and general vision knowledge from CLIP
\cite{radford2021learning}. The results in Table \ref{tab:result_dfer} indicate
that our method demonstrated significant improvements over these
vision-language models (+5.84\% UAR and +4.78\% WAR on DFEW, +2.70\% UAR and
+0.91\% WAR on FERV39K, +4.51\% UAR and +4.82\% WAR on MAFW).

Additionally, we present the detailed performance of each emotion on DFEW in
Table \ref{tab:result_dfew}. The results show that S2D outperforms the previous
best methods across most facial expressions, regardless of whether they are
supervised (e.g., IAL \cite{li2023intensity}, M3DFEL \cite{wang2023rethinking})
or self-supervised methods (e.g., MAE-DFER \cite{sun2023mae}). These
improvements are particularly evident in \emph{happy}, \emph{sad},
\emph{neutral}, \emph{angry}, and \emph{surprise}. For instance, S2D surpasses
M3DFEL by 14.87\%, 7.43\%, 9.95\% on \emph{sad}, \emph{neutral}, and
\emph{angry}, respectively. In contrast to MAE-DFER, which achieved an unbiased
representation through massive-scale self-supervised learning and performed
well on \emph{disgust}, supervised models exhibit bias in modeling
\emph{disgust} due to the lack of related training samples (occupying only
1.2\% of the entire dataset). To mitigate this, we trained the S2D model with
an oversampling strategy additionally, which greatly improves the performance
on the minor classes, \emph{disgust} and \emph{fear} (25.52\%$-$1.38\%=24.14\%,
50.22\%$-$34.71\%=15.51\%), while keeping the highest UAR and WAR metrics.

\emph{In summary, the promising results on three in-the-wild datasets demonstrate the strong generalization ability of S2D. We emphasize that our proposed method does not require retraining all model parameters on the DFER dataset. Instead, we only fine-tune a small number of parameters (less than 10\% of tunable parameters), making it more parameter-efficient and practical.}
\begin{table*}[htb]
	\centering
	\caption{Detailed comparison with state-of-the-art methods on DFEW as regards specific accuracy numbers of each emotion class. *: sampling two clips uniformly along the temporal axis for each video sample and then calculating the average score as the final prediction during testing time. ${ }^{\dagger}$: training with oversampling strategy.  UAR: unweighted average recall; WAR: weighted average recall.}
	\begin{tabular}{lcccccccccc}
		\toprule
		\multirow{2}{*}{Method}                           & \multirow{2}{*}{\tabincell{c}{Tunable                                                                                                                                                                                                 \\Params (M)}} & \multicolumn{7}{c}{Accuracy of Each Emotion (\%) } & \multicolumn{2}{c}{Metric (\%)} \\ \cmidrule(lr) { 3-9 } \cmidrule(lr) { 10-11 }
		                                                  &                                       & Happy                   & Sad                     & Neutral           & Angry             & Surprise          & Disgust           & Fear              & UAR               & WAR               \\
		\midrule
		\textit{\textbf{Supervised learning models}}      &                                       &                         &                         &                   &                   &                   &                                                                               \\

		C3D \cite{Tran2014LearningSF}                     & 78                                    & 75.17                   & 39.49                   & 55.11             & 62.49             & 45.00             &
		1.38                                              & 20.51                                 & 42.74                   & 53.54                                                                                                                                                               \\ R(2+1)D-18 \cite{tran2018closer} & 33    & 79.67 &
		39.07                                             & 57.66                                 & 50.39                   & 48.26                   & 3.45              & 21.06             & 42.79             & 53.22                                                                         \\ 3D ResNet-18
		\cite{He2015DeepRL}                               & 33                                    & 76.32                   & 50.21                   & 64.18             & 62.85             & 47.52             & 0.00              & 24.56
		                                                  & 46.52                                 & 58.27                                                                                                                                                                                         \\ ResNet18+LSTM \cite{Zhao2021FormerDFERDF} & -     & 83.56 &
		61.56                                             & 68.27                                 & 65.29                   & 51.26                   & 0.00              & 29.34             & 51.32             & 63.85                                                                         \\ Former-DFER
		\cite{Zhao2021FormerDFERDF} [MM'21]               & 18                                    & 84.05                   & 62.57                   & 67.52             & 70.03             &
		56.43                                             & 3.45                                  & 31.78                   & 53.69                   & 65.70                                                                                                                                     \\
		CEFLNet \cite{liu2022clip} [IS'2022]              & 13                                    & 84.00                   & 68.00                   & 67.00             & 70.00             &
		52.00                                             & 0.00                                  & 17.00                   & 51.14                   & 65.35                                                                                                                                     \\ EST \cite{liu2023expression} [PR'23] &
		43                                                & 86.87                                 & 66.58                   & 67.18                   & 71.84             & 47.53             & 5.52              & 28.49             & 53.43             & 65.85                                 \\
		STT \cite{Ma2022SpatioTemporalTF} [arXiv'22]      & -                                     & 87.36                   & 67.90                   & 64.97             &
		71.24                                             & 53.10                                 & 3.49                    & 34.04                   & 54.58             & 66.65                                                                                                                 \\ NR-DFERNet \cite{li2022nr}
		[arXiv'22]                                        & -                                     & 88.47                   & 64.84                   & 70.03             & 75.09             & 61.60             & 0.00              & 19.43             & 54.21             &
		68.19                                                                                                                                                                                                                                                                                     \\ NR-DFERNet${ }^{\dagger}$ \cite{li2022nr} [arXiv'22]                                               & -                   & 86.42 & 65.10 &
		70.40                           & 72.88               & 50.10 &
		0.00                            & 45.44              & 55.77 &
		68.01                                                                                                                                                                                                                                                                   \\ IAL \cite{li2023intensity} [AAAI'23] & 19    & 87.95 &
		67.21                                             & 70.10                                 & 76.06                   & 62.22                   & 0.00              & 26.44             & 55.71             & 69.24                                                                         \\ M3DFEL
		\cite{wang2023rethinking} [CVPR'23]               & -                                     & 89.59                   & 68.38                   & 67.88             & 74.24             & 59.69
		                                                  & 0.00                                  & 31.64                   & 56.10                   & 69.25                                                                                                                                     \\
		\textbf{S2D (ours)}                               & 9                                     & \underline{93.87}       & \textbf{83.25}          & \underline{75.31} & \textbf{84.19}    & \underline{64.33} & 0.00              & 37.07             & 62.57             & \underline{75.98} \\
		\textbf{S2D (ours)  * }                           & 9                                     & 93.62                   & \underline{80.25}       & \textbf{77.14}    & \underline{81.09} & \textbf{64.53}    & 1.38              & 34.71             & 61.82             & \textbf{76.03}    \\
		\textbf{S2D (ours)  *${ }^{\dagger}$ }            & 9                                     & \textbf{93.95}          & 78.35                   & 70.25             & 78.00             & 61.88             & \textbf{25.52}    & \textbf{50.22}    & \textbf{65.45}    & 74.81             \\
		\midrule
		\textit{\textbf{Self-supervised learning models}} &                                       &                         &                         &                   &                   &                   &                                                                               \\
		MAE-DFER \cite{sun2023mae} * [MM'23]              & 85                                    & 92.92                   & 77.46                   & 74.56             & 76.94             & 60.99             & \underline{18.62} & \underline{42.35} & \underline{63.41} & 74.43             \\
		\bottomrule
		\\
	\end{tabular}
	\label{tab:result_dfew}
\end{table*}

\subsection{Visualization Analysis}
\begin{figure*}[htb]

	\begin{flushright}
		\begin{overpic}[width=0.91\textwidth]{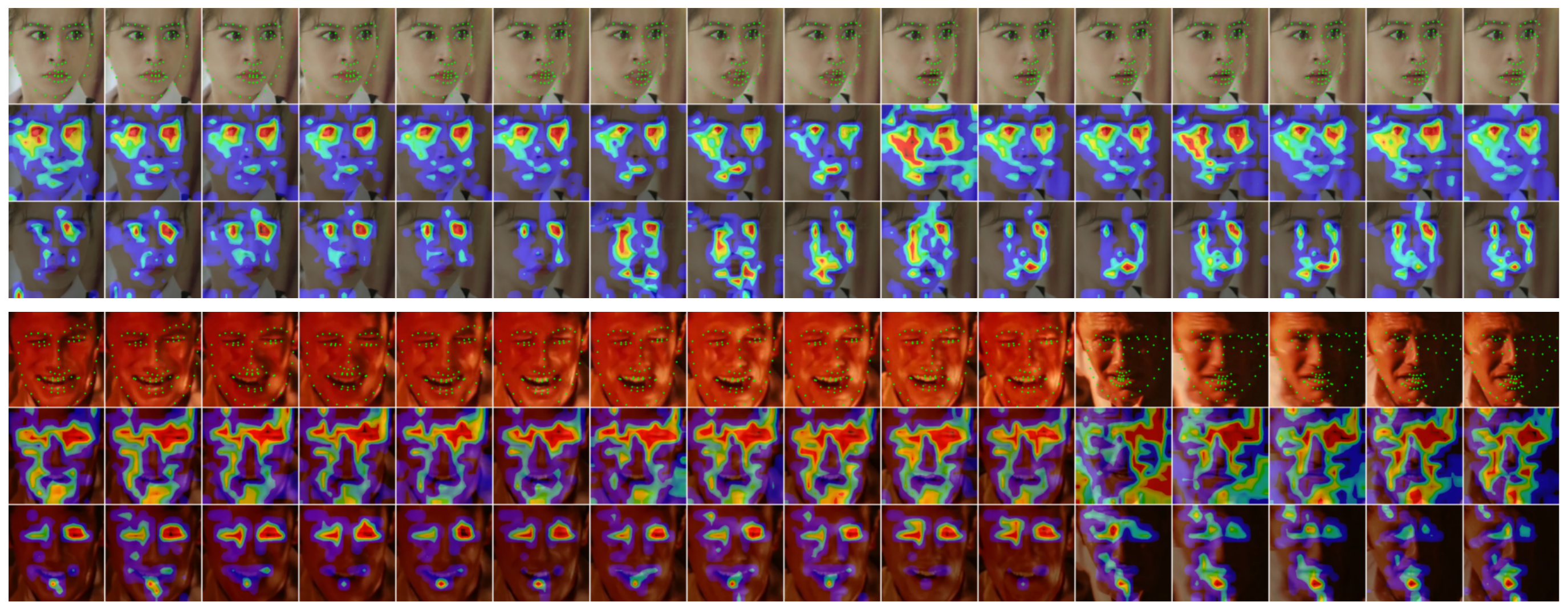}
			\put(-11.5,39){\small Frame number}
			\put(3,39) {\small 1 }
			\put(9.129999999999999,39) {\small 2 }
			\put(15.26,39) {\small 3 }
			\put(21.39,39) {\small 4 }
			\put(27.52,39) {\small 5 }
			\put(33.65,39) {\small 6 }
			\put(39.78,39) {\small 7 }
			\put(45.91,39) {\small 8 }
			\put(52.04,39) {\small 9 }
			\put(58.17,39) {\small 10 }
			\put(64.3,39) {\small 11 }
			\put(70.42999999999999,39) {\small 12 }
			\put(76.56,39) {\small 13 }
			\put(82.69,39) {\small 14 }
			\put(88.82,39) {\small 15 }
			\put(95,39) {\small 16 }
			\put(-8.5,34.5){\small input}
			\put(-8,29){\small w/o }
			\put(-10,27){\small landmarks}
			\put(-7.5,23){\small w/}
			\put(-10,21){\small landmarks}

			\put(-8.5,15.5){\small input}
			\put(-8,11){\small w/o }
			\put(-10,9){\small landmarks}
			\put(-7.5,4.5){\small w/}
			\put(-10,2.5){\small landmarks}

		\end{overpic}
	\end{flushright}
	\caption{Visualization of input facial image sequence with landmarks (row \textcolor{red}{1}), and attention scores (only positive values) of the last transformer block without and with facial landmark-aware features (row \textcolor{red}{2} and row \textcolor{red}{3} ). The darker the color, the greater the attention score. The expressions of the two sequences are \emph{surprise} and \emph{angry}, respectively. The S2D model with facial landmark knowledge is highly concentrated on the key regions of the face, while the model without facial landmarks is more likely to focus on the background and other irrelevant regions. The previous enhanced model's attention is more sensitive to changes in muscle movements while the latter consistently focuses on specific regions and remains relatively unchanged.}
	\captionsetup{justification=centering}
	\label{fig:visual}
\end{figure*}
\begin{figure*}[htp]
	\centering
	\begin{overpic}[width=1\textwidth]{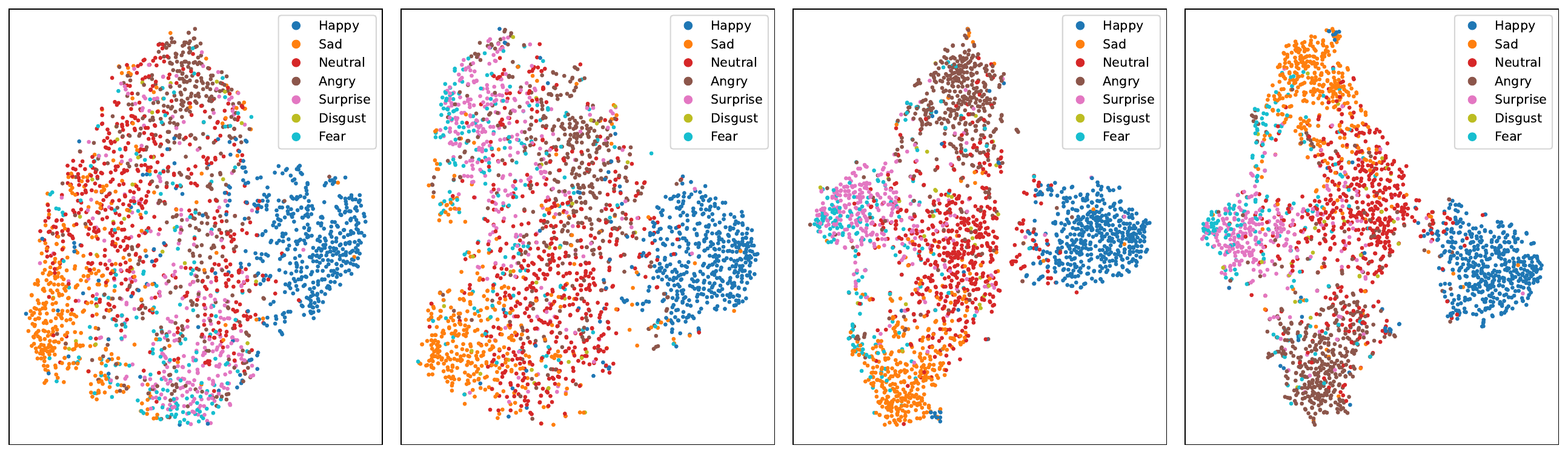}
		\put(9.5,29){\small Baseline}
		\put(32,29){\small Baseline+MCP}
		\put(54.5,29){\small Baseline+MCP+TMA}
		\put(81,29){\small Final S2D (Ours)}
	\end{overpic}
	\caption{High-level feature visualization on DFEW (fd1) using t-SNE \cite{van2008visualizing}. Here, we compare our final S2D model with baseline model, baseline+MCP model and baseline+MCP+TMA model.}
	\captionsetup{justification=centering}
	\label{fig:t-sne}
\end{figure*}
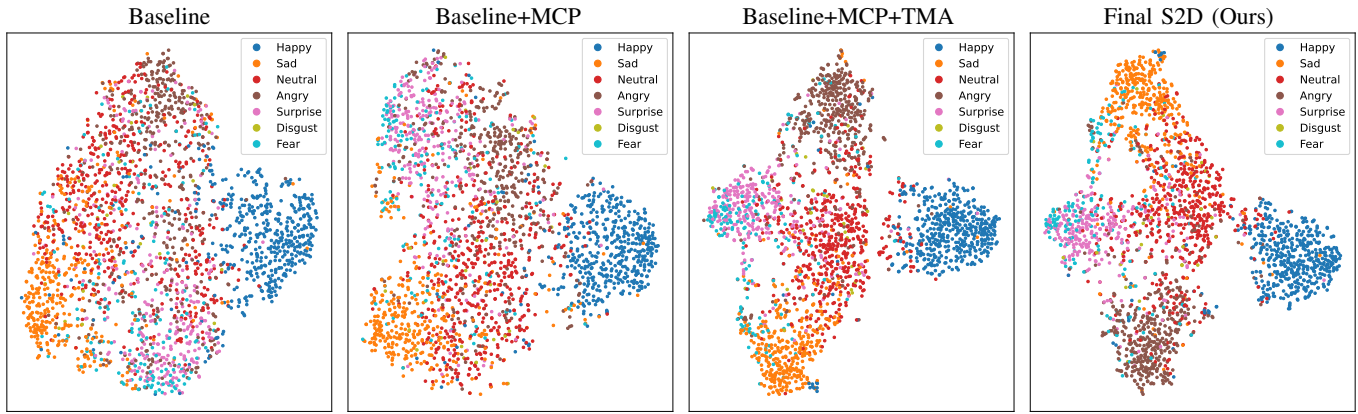

\subsubsection{Attention Visualization}
To confirm the critical role of the facial landmark-aware features in the S2D
model, we visualize the attention scores of the last transformer block in Fig.
\ref{fig:visual}. Our observations reveal that the attention of the S2D model
with facial landmark knowledge is highly concentrated on the key regions of the
face, which are more informative for FER. In contrast, the attention of the
model without facial landmark knowledge is more dispersed, suggesting a
tendency to focus on the background and other irrelevant regions. For example,
in the first sequence, frame 11 of row 3 demonstrates that the model's
attention is primarily directed towards the eyes and mouth, rather than other
irrelevant regions as shown in frame 11 of row 2. This distinction is even more
pronounced in the second sequence, particularly in frame 2 of row 3 compared to
frame 2 of row 2. These demonstrate that the facial landmark-aware feature
directs the model's attention towards emotion-related facial regions
successfully, thus benefiting FER.

Furthermore, by examining the changes in attention over time, we observed that
the model without landmark knowledge consistently focuses on specific regions
(e.g., eyes in the first sequence row 2, and both eyes and faces in the second
sequence row 2) and remains relatively unchanged. In contrast, the model with
landmark knowledge exhibits greater sensitivity to changes in muscle movements,
with its attention shifting accordingly in row 3 of the two sequences. For
example, in the first sequence, when the woman opens her mouth, the model's
attention is primarily on her lips, gradually diminishing as her mouth closes
(from frame 6 to frame 13). A similar phenomenon also can be observed in the
second sequence, where the model's attention initially focuses on the eyes and
mouth from frame 1 to frame 11 in row 3, then disappears on the mouth in frame
4, gradually weakens, and finally completely disappears on the mouth from frame
5 to frame 11. These observations strongly demonstrate that our model
accurately captures the dynamic changes of facial emotions in the temporal
dimension, with muscle movements implicitly encoded in facial landmark-aware
features.

\subsubsection{Visualization of Feature Distribution}
We gradually added MCP, TMA and SDL to the baseline model and visualized the
high-level features of the DFEW (fd 1) testing set using t-SNE
\cite{van2008visualizing}. Fig. \ref{fig:t-sne} illustrates that our proposed
modules significantly enhance the discriminative and separable qualities of
high-level features compared to the baseline model. The inclusion of the TMA
module notably improves the discriminability of high-level features, indicating
its effectiveness in capturing dynamic facial expression changes for DFER.
Compared to column 3, each category's feature distribution in column 4 is more
concentrated, particularly for \emph{happy}, which exhibits a more distinct
boundary than other expressions. This is attributable to the proposed SDL.
However, it is notable that the features corresponding to \emph{surprise} and
\emph{fear} are intermixed and challenging to distinguish. Additionally,
\emph{disgust} features are dispersed, lacking a clear clustering center. This
aligns with the emotion-specific accuracies in Table \ref{tab:result_dfew} and
is attributed to the data shortage of \emph{disgust} in the dataset available.

\begin{figure}
	\centering
	\begin{overpic}[width=0.49\textwidth]{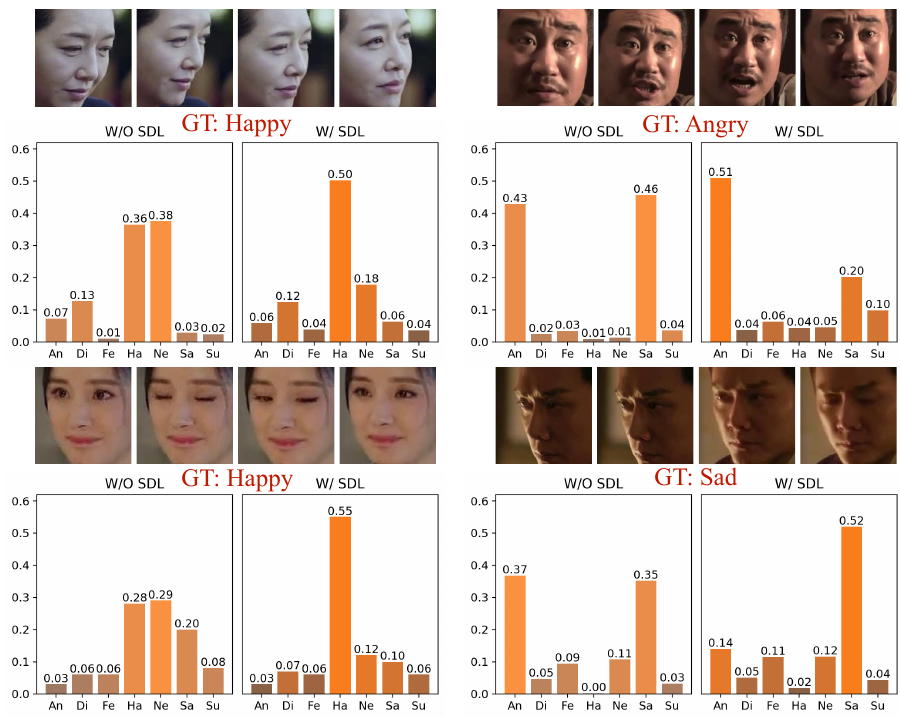}
	\end{overpic}
	\caption{Visualization of the output probability distribution.
		Here, \emph{An}, \emph{Di}, \emph{Fe}, \emph{Ha}, \emph{Ne}, \emph{Sa} and \emph{Su} denote \emph{angry}, \emph{disgust}, \emph{fear}, \emph{happy}, \emph{neutral}, \emph{sad} and \emph{surprise}, respectively. The model with SDL shows higher discriminability for ambiguous expressions. GT: ground truth.}
	\captionsetup{justification=centering}
	\label{fig:prob}
\end{figure}
\subsubsection{Visualization of Output Probability Distribution}
To evaluate the effectiveness of our proposed SDL, a comparison was conducted
between the output probability distributions of the S2D model with and without
SDL. As illustrated in Fig. \ref{fig:prob}, the woman in the top left corner
displays a subtle smile, challenging to discern due to minimal changes in her
facial muscles. The model without SDL assigns high scores to both \emph{happy}
and \emph{neutral} (depicted in the left bar chart) with a marginal difference
of 0.02, suggesting the prediction is ambiguous. Conversely, the SDL-enhanced
S2D model assigns a higher score to \emph{happy} than \emph{neutral} by 0.23
(right bar chart). This phenomenon is similarly observed in the bottom-left
example. In the right column of Fig. \ref{fig:prob}, although the \emph{angry}
and \emph{sad} are clearly expressed on the faces of the two men, the model
without SDL provides similar scores for the \emph{angry} and \emph{sad} (0.43
vs. 0.46, 0.37 vs. 0.35), resulting in wrong predictions. In contrast, the S2D
equipped with SDL effectively distinguishes \emph{angry} and \emph{sad} with
high discriminability.

The soft labels provided by SDL have fewer ambiguities and help reduce the
model's uncertainty. Consequently, the SDL-incorporated model demonstrates
higher discriminability for ambiguous expressions (e.g., smile and neutral) and
the model without SDL struggles to distinguish between certain ambiguous
samples (e.g., right examples in Fig. \ref{fig:prob}).

\section{Conclusion}
In this paper, we propose a simple yet powerful framework, S2D, which adapts a
landmark-aware image model for facial expression recognition in videos. This
study indicates that the prior knowledge from SFER data and facial landmark
detections can be leveraged to enhance DFER performance. The Multi-View
Complementary Prompters (MCPs) employed in this work effectively utilize static
facial expression features learned on AffectNet
\cite{Mollahosseini2017AffectNetAD} dataset and facial landmark-aware features
from MobileFaceNet \cite{Chen2018MobileFaceNetsEC}. Furthermore, the S2D is
extended from the static model efficiently with Temporal-Modeling Adapters
(TMAs) and significantly enhanced by our Emotion-Anchors based
Self-Distillation Loss (SDL). Experimental results on widely used benchmarks
consistently reveal that S2D achieves performance comparable to previous
state-of-the-art methods, demonstrating our model's ability to learn robust
image-level representations for SFER and powerful dynamic facial
representations for DFER. Lastly, compared to previous self-supervised methods,
S2D is much more parameter-efficient and practical for DFER, with only 9M
tunable parameters. We believe it will serve as a solid baseline and contribute
to relevant research. In the future, we will explore more effective ways to
leverage prior facial knowledge and other potential knowledge in the SFER model
for improving DFER performance.

\bibliographystyle{IEEEtran}
\bibliography{egbib}

\vfill

\end{document}